\documentclass[pdflatex,sn-mathphys]{sn-jnl}
\usepackage{graphicx}
\usepackage{epsfig}
\usepackage{amssymb}
\usepackage{mathrsfs}
\usepackage{bm}
\usepackage{color}
\usepackage{amsmath}
\usepackage{mathrsfs}
\usepackage{algorithm}  
\usepackage{algorithmicx}  
\usepackage{algpseudocode}  
\usepackage{booktabs}
\usepackage{multirow}
\usepackage{float}

\jyear{2021}%

\theoremstyle{thmstyleone}%
\theoremstyle{thmstyletwo}%

\theoremstyle{thmstylethree}%

\begin{document}

\title[A Missing Value Filling Model Based on Feature Fusion Enhanced Autoencoder]{A Missing Value Filling Model Based on Feature Fusion Enhanced Autoencoder}


\author{\fnm{Xinyao} \sur{LIU}}\email{liuxinyao@my.swjtu.edu.cn}
\author*{\fnm{*Shengdong} \sur{DU}}\email{sddu@swjtu.edu.cn}
\author{\fnm{Tianrui} \sur{LI}}\email{trli@swjtu.edu.cn}
\author{\fnm{Fei} \sur{TENG}}\email{fteng@swjtu.edu.cn}
\author{\fnm{Yan} \sur{YANG}}\email{yyang@swjtu.edu.cn}

\affil{\orgdiv{School of Computing and Artificial Intelligence}, \orgname{Southwest Jiaotong University}, \orgaddress{\city{Chengdu}, \postcode{611756}, \country{P.R.China}}}


\abstract{With the advent of the big data era, the data quality problem is becoming more critical. Among many factors, data with missing values is one primary issue, and thus developing effective imputation models is a key topic in the research community.
Recently, a major research direction is to employ neural network models such as self-organizing mappings or automatic encoders for filling missing values.
However, these classical methods can hardly discover interrelated features and common features simultaneously among data attributes. Especially, it is a very typical problem for classical autoencoders that they often learn invalid constant mappings, which dramatically hurts the filling performance.
To solve the above-mentioned problems, we propose a missing-value-filling model based on a feature-fusion-enhanced autoencoder. 
We first incorporate into an autoencoder a hidden layer that consists of de-tracking neurons and radial basis function neurons, which can enhance the ability of learning interrelated features and common features. Besides, we develop a missing value filling strategy based on dynamic clustering that is incorporated into an iterative optimization process. This design can enhance the multi-dimensional feature fusion ability and thus improves the dynamic collaborative missing-value-filling performance.
The effectiveness of the proposed model is validated by extensive experiments compared to a variety of baseline methods on thirteen data sets.
}



\maketitle

\section{Introduction}\label{sec1}

Data quality issue is one of the key challenges in many research fields such as data science, data mining, and machine learning. Good-enough data quality is often a prerequisite to many downstream tasks. If this issue is not handled properly, unexpected outcomes or even wrong conclusions can be drawn, which is known as the \textquotedblleft Garbage In Garbage Out\textquotedblright\ problem \cite{bib1}.
Among many factors, data containing missing values is one primary reason that harms data quality and has received enormous attention from the research community. Many real-world applications do generate incomplete or fragmented data pieces \cite{20} but most learning models do not readily work with missing values. Therefore, many researchers have proposed and developed various missing-value-filling methods from different perspectives \cite{bib3}\cite{5}\cite{8}. The principle is to properly estimate the distribution of missing values and then induce deterministic numbers from the uncertainty.

\par While many missing value estimations are based on conventional statistical models, the recent advances of deep learning models have provided a new perspective \cite{bib2}. Deep neural nets excel in in non-linearity approximation, and even models with simple architectures can achieve decent performance. Example models are Self-Organizing Map (SOM) \cite{2}, Multi-Layer Perceptron (MLP) , and AutoEncoder (AE) \cite{4}. On the other hand, the functionality of simple models is also limited. For example, the SOM model ignores the correlation among data attributes, thus resulting in low model accuracy \cite{bib3}; the MLP-based model training process is overly time-consuming because of its high computational complexity; the AE-based model, which is most related to our work, has effectively reduced the model complexity by implementing only one network structure, but the model outputs are likely to track the corresponding inputs, thus leading to an invalid identity mapping learned and showing the self-tracking problem \cite{5}. 
Hence, many improvements have been made upon the basic autoencoder-based model. For instance, Radial Basis Function Neural Network (RBFNN) \cite{6}, Generalized Regression Neural Network (GRNN) \cite{bib9}, Tracking-removed Autoencoder (TRAE)\cite{5}, and Correlation-enhanced Auto-associative Neural Network (CE-AANN)\cite{8}. However, the above methods could hardly learn the common features and the interrelated features simultaneously. Especially when using a only basic autoencoder to fill missing values, it often learns invalid constant mappings.\footnote{This paper is an extended version of \cite{19}, which has been accepted for presentation at the 15th International FLINS Conferences on Machine learning, Multi agent and Cyber physical systems (FLINS2022).}

\par To address the aforementioned problems, this paper proposes a feature-fusion-enhanced and autoencoder-based missing-value-imputation model (FFEAM). The first novelty is that we incorporate a hidden layer comprised of de-tracking neurons and radial basis function neurons, which can enhance the multi-dimensional feature fusion ability. 
De-tracking neurons can avoid learning invalid constant mappings that often occur in classical autoencoders; besides, the neurons can also explore the interrelated features among data attributes effectively.
Radial basis function neurons are designed to have an automatic clustering ability, which can better learn the common features among data missing samples. 
The outputs of these two types of neurons are further constrained by each other, and thus the model can simultaneously discover common features and interrelated features among data attributes. 
The second novelty is that we develop a missing value filling strategy based on dynamic clustering (MVDC) that is incorporated into an iterative optimization process. Specifically, the MVDC strategy clusters the current training data and feeds the selected centroids and widths to the radial basis function neurons, while the missing values are considered as variables along with the model parameters and are jointly optimised.
As the optimization proceeds, the estimation of the missing values will be more accurate, and thus the filling precision will gradually improve.

\par The main contributions of this paper can be summarized as follows:
\begin{itemize}
    \item We propose a missing value filling model based on a feature-fusion-enhanced autoencoder. We address the typical problem occurring in classical autoencoders by adding a hidden layer that consists of de-tracking neurons and radial basis function neurons. By exchanging information and constraining each other's outputs, both types of neurons enable the model to effectively learn interrelated features and common features across different data dimensions.
    \item We develop a missing value filling strategy based on dynamic clustering that is incorporated into an iterative optimization process. This design can enhance the multi-dimensional feature fusion ability and thus improves the dynamic collaborative missing-value-filling performance.
    \item We conducted extensive experiments on seven publicly available datasets and six artificial datasets, of which the results demonstrate that the proposed model achieves a better performance compared to many benchmark models under different missing value conditions.
\end{itemize}

\par The rest of this paper is organized as follows. Section \ref{s2.0} introduces the related work. Section \ref{s3.0} first presents the preliminaries and then details the model implementation. Section \ref{s4.0} introduces the architecture of FFEAM. Section \ref{s5.0} validates the effectiveness of the proposed model through comparative experiments, and finally, Section \ref{s6.0} concludes this paper.

\section{Related Work}\label{s2.0}

The current popular missing value filling methods can be divided into two categories: statistical models and machine learning models \cite{bib4}. For the first category, the mean filling method is utilized as the classical method which primarily involves utilizing the average value of a data attribute column to fill in missing values \cite{9}. The Expectation-Maximization (EM) filling method uses the marginal distribution of the available data to perform the Maximum Likelihood Estimate (MLE) on the missing data to analyze the most likely value to be obtained for the missing value filling \cite{10}. In the research of missing value filling models based on machine learning, K-Nearest Neighbors (KNN) is one of the classical methods, and its main idea is to select the top K categories with the least variance by sorting the variance between the categorical samples and the training samples from smallest to largest \cite{12}. Tutz et al. proposed a weight parameter adjustment method to optimize the weight between samples under different K values to improve the effect of missing value filling \cite{13}. The ordered nearest neighbor imputation method is an enhanced version of the KNN imputation, which utilizes filled data samples from prior missing data to improve data utilization rates \cite{14}. Migdady et al. proposed an enhanced fuzzy k-means clustering approach that employs a K-Means clustering model to partition the input dataset into K clusters \cite{15}. Li et al. proposed a missing value filling method based on spatio-temporal multi-view learning, which automatically fill missing records of geosensing data by learning from multiple views from both local and global perspectives \cite{16}. Deng et al. proposed an improved random forest padding algorithm, which combines linear interpolation, matrix combination and matrix transposition, to solve the padding problem \cite{26}. Noei et al, proposed a hybrid GA and ARO algorithm (GARO), which significantly improves the computation time of genetic algorithms used to interpolate missing values \cite{27}. Mostafa et al. presented two interpolation methods that utilize Bayesian Ridge techniques for feature selection under two different conditions \cite{28}.

\par Since the deep learning method was proposed in 2006, it has attracted the attention of researchers \cite{bib5}. The deep learning model can automatically extract and learn the deep features in the data through the multi-layer neural network \cite{21}. Currently, deep learning-based missing value filling methods are becoming a hot research topic \cite{bib6}. For instance, Ravi et al proposed more variants of classical autoencoder models  for missing value filling, and the experiments demonstrated that the generalized regression autoencoder has better filling performance in the family of autoencoder-based architectures \cite{bib9}. However, the generalized regression autoencoder has high computational complexity due to the need to repeatedly calculate the distance between incomplete data and all complete data during filling. Considering that the classical autoencoder has a certain dependence on the model input and may learn invalid constant mappings. In order to reasonably weaken the tracking of the input by the autoencoder, Lai et al. developed a dynamic filling method for the de-tracking autoencoder, whose hidden layer neurons can avoid the self-tracking of the network output to the corresponding input by dynamically organizing the input structure \cite{5}. Lai et al. also continued their improvement to propose the association enhanced autoassociative neural network model for missing value filling, which can well explore the association features among data attributes \cite{8}. Wang et al. developed a new model based on gated recurrent units (GRU), which can efficiently learn the internal features of IoT data as well as the historical information of time series data \cite{25}.
\par The above methods such as mean filling, K-nearest neighbor filling \cite{22}, and clustering-based filling \cite{15} mostly analyze and process the missing data from the data common feature perspective, i.e., fill the data by learning the similarity features between data. Due to its nonlinear feature learning ability, deep neural network can automatically mine complex nonlinear relationships between data attributes, and can learn the interrelated features between data attributes to fill in missing values. However, it is difficult for the proposed missing value filling models to perform feature fusion learning from the above two dimensions (common features and interrelated features) at the same time.

\section{Preliminaries}\label{s3.0}
\subsection{Classical autoencoder}

Autoencoder(AE) is a classical deep learning model consisting of an encoder and decoder whose output attempts to reconstruct its corresponding input\cite{18}. Since the autoencoder can reproduce the value of the input, if there are missing values in the sample, it can be filled by the value reproduced at the output. The model achieves input reconstruction and missing value filling by minimizing the cost function shown in Equation (1).

\begin{equation}
L = \frac{{\rm{1}}}{{{\rm{2}}n}}\sum\limits_{i = 1}^n {\sum\limits_{j = 1}^{\rm{s}} {{{({y_{ij}} - {x_{ij}})}^2}} },\label{pythagorean}
\end{equation}
where $n$ represents the number of training samples; $s$ represents the number of attributes of the samples; $x_{ij}$ represents the input values, i.e., the data containing pre-filled missing values; and $y_{ij}$ represents the filled values of the model. To continuously enhance the model, efforts are made to minimize training errors. However, this may lead to reconstructed outputs that become invalid due to constant mappings, resulting in high similarity or equality between model inputs and outputs. These outputs tend to closely track the corresponding inputs, thereby presenting a self-tracking problem.

\subsection{Radial basis function autoencoder }
The radial basis function autoencoder consists of an input layer, a single hidden layer, and an output layer \cite{6}. In this model, the connection weight between the input layer and the hidden layer is fixed to 1, while the connection weight between the hidden layer and the output layer is used as a parameter for the training of the model, and the computation process of the radial basis function hidden layer neurons is shown in Equation (2).

\begin{equation}
net_{ik}^{{\rm{(1)}}} = \exp \left[ { - \frac{{\left\| {{x_i} - \left. {{\mu _k}} \right\|} \right.}}{{2\sigma _k^2}}} \right],k = 1,2,...,{n^{(1)}},
\label{pythagorean}
\end{equation}
where $\mu _k$ represents the centroid of the $k$ hidden layer neurons; $\sigma_k$ is the width of the $k$ hidden layer neurons, which determines the magnitude of the decay of the function taking values along the center to the surrounding. According to the properties of radial basis function, when the input sample is far from the centroid, the activation of the neuron is approximately 0, which belongs to the neuron with low activation.  And when the input sample is closer to the center, the activation of the neuron is approximately 1, which belongs to the neuron with high activation. The output layer of the model uses a linear activation function, so the output of the model is approximately equal to the weighted sum of neurons with high activation. The radial basis function autoencoder model has a certain clustering ability, which can explore the common features of data and reduce the self-tracking of the classical autoencoder to a certain extent, but the model is weak in learning the interrelated features of data.

\subsection{Correlation-enhanced autoassociative neural network}
The hidden layer of the the correlation-enhanced autoassociative neural network model(CE-AANN) \cite{8} consists of $m_1$ traditional hidden layer neurons and $m_2$ improved hidden layer neurons and produces two outputs in the output layer for both types of neurons. Taking the $k$th hidden layer neuron as an example, the traditionally hidden layer neuron is solved according to Equation (3).

\begin{equation}
net_{ikj}^{(1)} = \phi (\sum\limits_{l = 1}^s {w_{lk}^{(1)}}  \cdot {x_{il}} + b_k^{(1)}),j = 1,2,...,s,
\label{pythagorean}
\end{equation}
where $\phi()$ denotes the activation function of the hidden layer neuron; $k$ represents the $k$th traditional hidden layer neuron; and $s$ represents the number of attributes.
The improved hidden layer neuron is solved according to Equation (4).

\begin{equation}
net_{ikj}^{(1)} = \phi (\sum\limits_{l = 1,l \ne j}^s {w_{lk}^{(1)}}  \cdot {x_{il}} + b_k^{(1)}),j = 1,2,...,s,
\label{pythagorean}
\end{equation}
where $\phi()$  denotes the activation function of the hidden layer neuron; $k$ represents the $k$th improved hidden layer neurons, and $s$ represents the number of attributes. From Equation (4), it can be seen that for the $j$th neuron in the output layer of the network, the hidden layer neuron dynamically rejects the value of $x_{ij}$, i.e., the corresponding input of $y_{ij}$ is rejected, and the input values other than $x_{ij}$ are used to solve the hidden layer output.
\par The cost function of the CE-AANN is shown in Equation (5).

\begin{equation}
L = \frac{{\rm{1}}}{{\rm{2}}}\sum\limits_{i = 1}^n {\sum\limits_{j = 1}^s [ } {({y_{ij}} - {x_{ij}})^2} + {({y_{ij}} - {r_{ij}})^2}],
\label{pythagorean}
\end{equation}
where $y_i$= [$y_{i1}$,$y_{i2}$,...,$y_{is}$]$^T$ denotes the network output; $r_i$= [$r_{i1}$,$r_{i2}$,...,$r_{is}$]$^T$ is the reference output; $n$ represents the total number of data; $s$ represents the number of sample attributes. According to Equation (5),it can be seen that the CE-AANN model constantly minimizes the error between the output $y_{ij}$  and input $x_{ij}$, and also de-tracked the similarity of input and output, through the mutual constraint of the two types of outputs, the model can weaken the dependence of the output on the input while making the network input have a certain borrowing effect on the output, so as to improve the model’s learning ability of data interrelated features, but the model is weak in learning common features of the data.

\section{Method}\label{s4.0}
\subsection{Overall framework of the model}
\begin{figure}[!htp]
\begin{center}
\includegraphics*[width=0.9\textwidth]{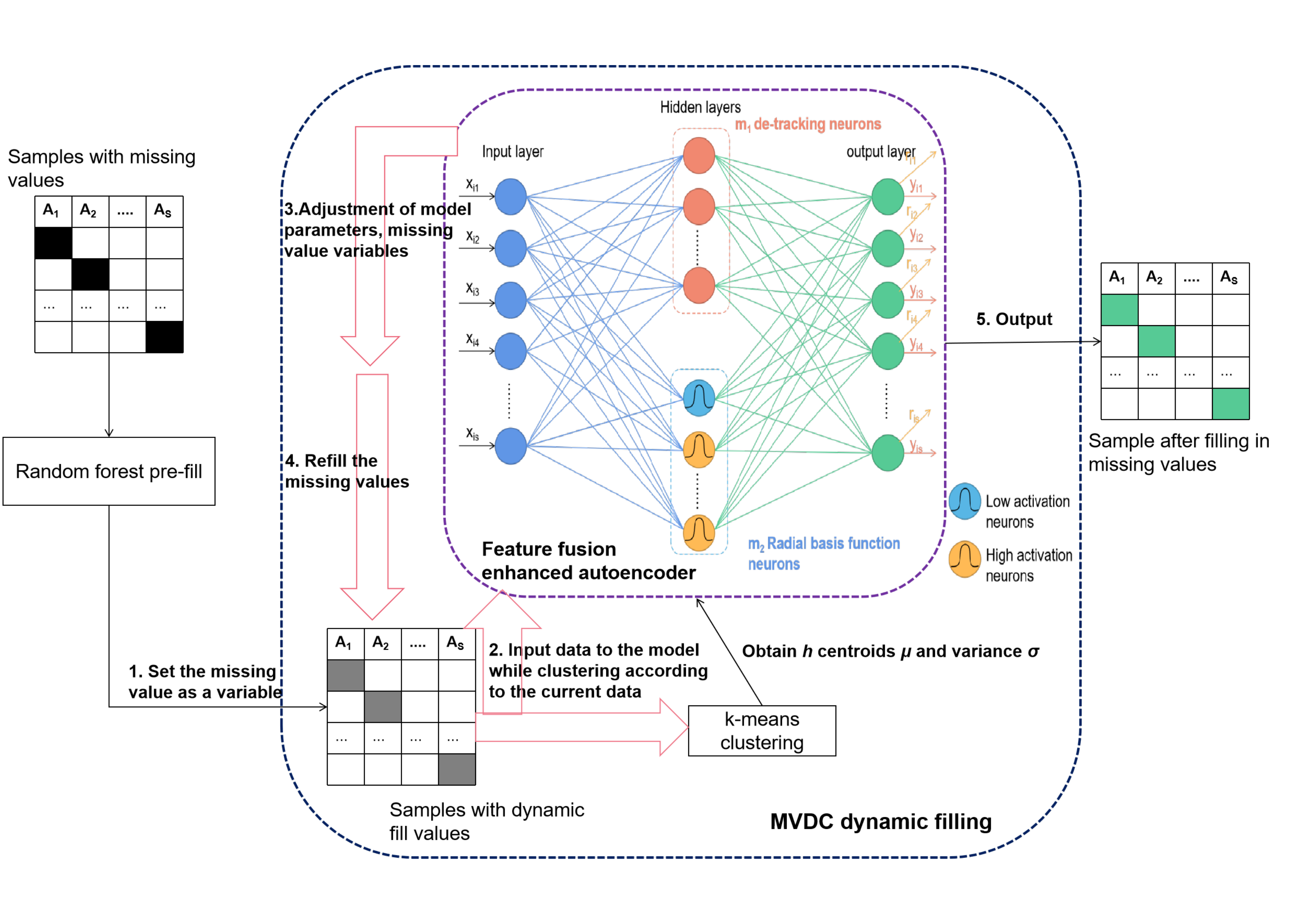}
\end{center}
\caption{Design diagram of Feature Fusion Enhanced Autoencoder Model for Missing Value Filling (FFEAM)}
\label{fig0}
\end{figure}

To address the main problems in the classical autoencoder model used for missing value filling, a missing value filling model based on feature fusion enhanced autoencoder is developed, and a novel neural network hidden layer based on de-tracking neurons and radial basis function neurons are designed to collaboratively train to fill missing values. Figure \ref{fig0} shows the overall framework of the model, which first pre-fills the missing data set by the random forest. The missing values are set as variables by the MVDC filling strategy, and then the samples containing dynamic filling values are input to the FFEAM, which performs two dimensions of fusion learning on the interrelated features and common features in the data. Meanwhile, the MVDC filling strategy, through K-means clustering self-organized learning, selects the centroids and variances required for the radial basis function neurons, and finally optimizes the missing filling effect iteratively.

\subsection{Feature Fusion Enhanced Autoencoder Design}
\begin{figure}[!htp]
\begin{center}
\includegraphics*[width=0.8\textwidth]{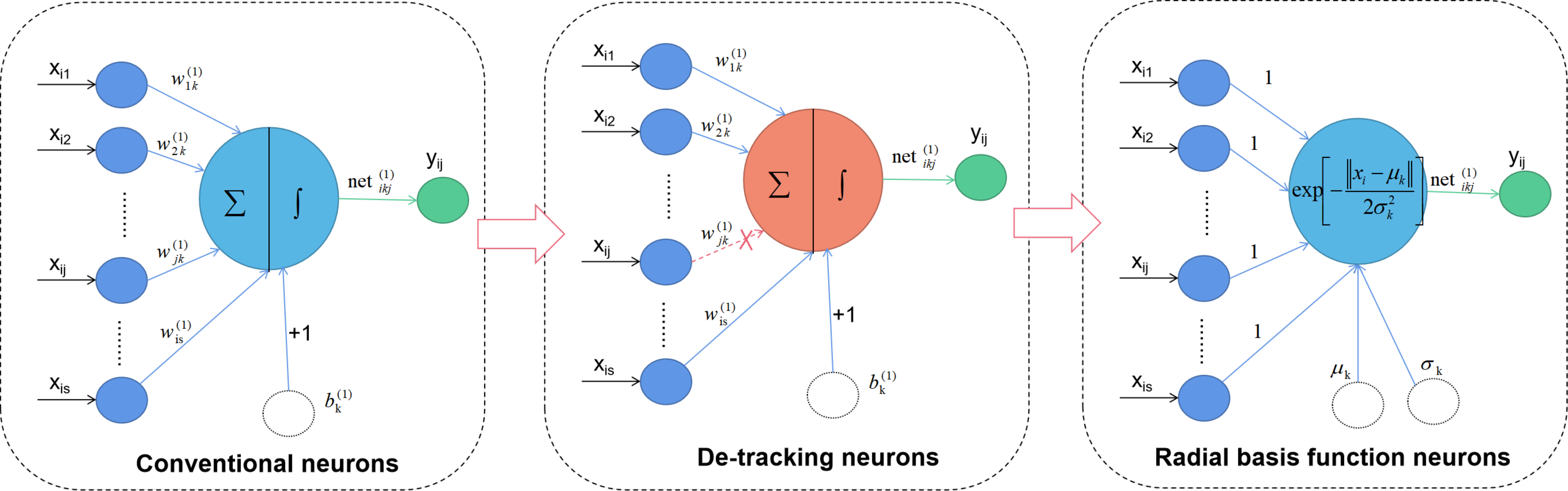}
\end{center}
\caption{Three different hidden layer neurons}
\label{fig1}
\end{figure}

As shown in Figure \ref{fig0}, the FFEAM contains an input layer, a modified novel hidden layer, and an output layer. in which two types of neurons are developed and jointly constructed: include $m_1$  de-tracking neurons and $m_2$ radial basis function neurons. According to the mutual design of the two types of hidden layer neurons, two types of outputs will be generated in the output layer: $y_i$= [$y_{i1}$,$y_{i2}$,...,$y_{is}$]$^T$  represents the output of the network based on the solution of the de-tracking hidden layer neurons, $r_i$= [$r_{i1}$,$r_{i2}$,...,$r_{is}$]$^T$ is the reference output based on the radial basis function hidden layer neuron solution. To introduce the design idea of the novel hidden layer in detail, Figure \ref{fig1} shows the difference between the conventional neuron, the de-tracking neuron, and the radial basis function neuron with the $k$th hidden layer neuron as an example.

As shown in Figure \ref{fig1}, the output of a conventional neuron is typically computed using Equation (1), and such hidden layer neurons are frequently employed in self-encoders. However, one drawback is that when using these neurons in self-encoders, it may be easy to learn an output that simply replicates the input mapping, which can't effectively extracting interrelated features and common features of the data. On the other hand, de-tracking neuron calculate their output using Equation (4). Figure \ref{fig1} shows how these neurons dynamically discard $x_{ij}$ based on the output $y_{ij}$. This ensures that the self-encoder does not learn invalid mappings, which would lead to directly reproducing the input. As a result, de-tracking neurons outperform traditional neurons in extracting inter-attribute interrelated features in the data. Nonetheless, they face the same limitation as traditional neurons in discovering common features of the data attributes. Meanwhile, the output of the radial basis function neuron is calculated according to Equation (2), with lower activation as the input sample is further away from the centroid and higher activation as it gets closer, which allows the output of the radial basis function neuron to approximately represent a weighted sum of highly activated neurons, helping to cluster the data and discover common features. However, radial basis function neurons have limited ability to reveal interrelated features between data attributes.

\par The FFEAM model comprehensively considers the advantages and disadvantages of various types of neurons and complements the advantages of the two types of neurons through the fusion design of de-tracking neurons and radial basis function neurons. The de-tracking neuron reduces the self-tracking of the classical autoencoder by dynamically eliminating some inputs, and improves the model’s learning ability of data interrelated features. Moreover, the radial basis function neuron is activated by the radial basis function, and the network reference output can be approximately regarded as the weighted summation of the high activation neurons, which has a certain ability for cluster analysis and exploring the common features of the data. The multi-dimensional feature fusion learning is performed through the complementation of the above two types of neurons, thereby improving the missing value filling performance.
\par The whole FFEAM calculation process is described below, the input dataset of the model containing missing values 
$\{$$X_{ij}$$\|$$i$$=1,2,...,$$n$$;$$j$$=1,2,...,$$s$$\}$, where $n$ is the number of samples and $s$ is the number of attributes, and the missing values $X_{ij}$ are pre-filled using random forest. The method first traverses all the features and starts filling from the column with the least missing. After each regression prediction is completed, the predicted value is put back into the original feature matrix, and then the next feature is filled, and the pre-filling of all missing columns is completed in turn. Then the weights and thresholds of the FFEAM are initialized. Meanwhile, based on the MVDC dynamic filling strategy, the K-means clustering algorithm is used to find $h$ centroids $\mu_h$ in the data set after pre-filling, and the width $\sigma_g$ is calculated by Equation (6).

\begin{equation}
{\sigma _g} = \frac{{{c_{max}}}}{{\sqrt {2h} }}(g = 1,2,....,h),
\label{pythagorean}
\end{equation}
where $c_{max}$ denotes the maximum distance between $h$ centers; $h$ represents the number of centers and also the number of radial basis function neurons.
\par Next, the data association features are discovered using the de-tracking neurons in the novel hidden layer, and the data common features are discovered using the radial basis function neurons. The output of the de-tracking neuron is shown in Equation (7).
\begin{equation}
ne{t_{ikj}} = {\rm{relu (}}\sum\limits_{l = 1,l \ne j}^s {w_{lk}^{(1)}}  \cdot {x_{il}} + b_k^{(1)}),j = 1,2,...,s,k = 1,2...,{m_{\rm{1}}},
\label{pythagorean}
\end{equation}
where $ne{t_{ikj}}$ represents the output of the $k$th de-tracking neuron after eliminating the corresponding input $x_{ij}$, $s$ represents the number of attributes which is the number of columns in the $x_{ij}$ data set, $k$ represents the $k$th de-tracking neuron, $m_1$ is the number of de-tracking neurons, ${w_{lk}}^{(1)}$ represents the connection weight of the $l$th node in the input layer and the $k$th de-tracking neuron in the hidden layer, ${b_k}^{(1)}$ denotes the threshold of the $k$th de-tracking neuron of the hidden layer; from Equation (7), we can see that for the $j$th neuron of the output layer, the hidden layer neuron will dynamically reject $x_{ij}$ and use the input values other than  $x_{ij}$ to solve the hidden layer output.
\par The output of the radial basis function hidden layer neuron is shown in Equation (8).

\begin{equation}
ne{t_{igj}} = \exp \left[ { - \frac{{\left\| {{x_{ij}} - \left. {{\mu _g}} \right\|} \right.}}{{2\sigma _g^2}}} \right],j = 1,2,...,s,g = 1,2,...,{m_{\rm{2}}},
\label{pythagorean}
\end{equation}
where $ne{t_{igj}}$ represents the output of the $g$th radial basis function neuron for input $x_{ij}$, $g$ represents the $g$th radial basis function neuron, $s$ represents the number of attributes, and $m_2$ is the number of radial basis function neurons;  $\mu _g$ is the centroid of the g radial basis function hidden layer neurons, which is found according to the k-means algorithm. $\sigma_g$ is the width of the $g$ radial basis function hidden layer neurons, which determine the magnitude of the decay of the function taking values along the center to the surrounding, and is found according to Equation (6).
\par After the new hidden layer, two types of outputs are obtained in the output layer of the neural network, one is the output $y_{ij}$ of the de-tracking neuron in the corresponding hidden layer, which is calculated according to Equation (9).

\begin{equation}
y_{ij}^{} = \sum\limits_{k = 1}^{{m_{\rm{1}}}} {w_{kj}^{(2)}} ne{t_{ikj}} + b_j^{(2)},j = 1,2,...,s,
\label{pythagorean}
\end{equation}
where $ne{t_{ikj}}$ is the output of the $k$th de-tracking neuron; $s$ is the number of attributes; $m_1$ is the number of de-tracking neurons, ${w_{kj}}^{(2)}$ represents the connection weight of the $k$th de-tracking neuron in the hidden layer and the $j$th output layer neuron in the output layer, and ${b_j}^{(2)}$ represents the threshold value between the $j$th output layer neurons.
\par Second, the network output $r_{ij}$ corresponding to the radial basis function neuron in the hidden layer is calculated according to $Equation$ $(10)$.

\begin{equation}
r_{ij}^{} = \sum\limits_{g = {\rm{1}}}^{{m_{\rm{2}}}} {w_{gj}^{{\rm{(2)}}}} ne{t_{igj}} + b_j^{(2)},j = 1,2,...,s,
\label{pythagorean}
\end{equation}
where $ne{t_{igj}}$ is the output of the $g$th radial basis function neuron: $s$ is the number of attributes; $m_2$ is the number of radial basis function neurons; $w_{gj}^{(2)}$ represents the connection weight of the $g$th radial basis function neuron in the hidden layer and the $j$th output layer neuron in the output layer, and ${b_j}^{(2)}$ represents the threshold value between the $j$th output layer neurons. According to the activation function of Equation (8), when the input sample is far from the centroid, the activation of the neuron is approximately 0, which belongs to the neuron with low activation. And when the input samples are closer to the centroid, the neuron activation is approximately 1, which belongs to the neuron with high activation.
\par The model loss function is as follows.

\begin{equation}
L = \frac{{\rm{1}}}{{\rm{2}}}\sum\limits_{i = 1}^n {\sum\limits_{j = 1}^s [ } {({y_{ij}} - {x_{ij}})^2} + {({y_{ij}} - {r_{ij}})^2}],
\label{pythagorean}
\end{equation}

Through the above computational process, the model minimizes the error between the output of the de-tracking neuron $y_{ij}$ and the input $x_{ij}$ while being as close as possible to the reference output of the radial basis function neuron $r_{ij}$. Through the mutual constraint of the two types of outputs, the model is able to effectively learn the interrelated features and common features in the sample data while weakening the correlation between the inputs and outputs to reduce the self-tracking problem.

\subsection{Description of the algorithm process}
\begin{algorithm}[H]
  \caption{FFEAM algorithm}
     \begin{algorithmic}[1] 
       \Require 
       Data sets containing missing values: $\{$$X_{ij}$$\|$$i$$=1,2,...,$$n$$;$$j$$=1,2,...,$$s$$\}$.
       \Ensure 
       Filling the complete data set  $\{$$Y_{ij}$$\|$$i$$=1,2,...,$$n$$;$$j$$=1,2,...,$$s$$\}$.
       \State Pre-filling of missing values $X_{ij}$ using random forest;
       \State Set the missing values in $X_{ij}$ as variables;
       \State Initialize the weights $w_k$ , thresholds $b_k$,  and missing value variables;
       \State A number of centroids $\mu_h$ are found in $X_{ij}$ based on the k-means clustering algorithm, and the width $\sigma_g$ is calculated by equation (6);
       \Repeat
         \State Select the batch sample $X_b$ from the pre-filled sample of missing values;
          \Repeat
          \State Take a sample $x_{ij}$ in $X_b$;
           \State $ne{t_{ikj}} = {\rm{relu (}}\sum\limits_{l = 1,l \ne j}^s {w_{lk}^{(1)}}  \cdot {x_{il}} + b_k^{(1)}),j = 1,2,...,s,k = 1,2...,{m_{\rm{1}}}$;
            \State $ne{t_{igj}} = \exp \left[ { - \frac{{\left\| {{x_{ij}} - \left. {{\mu _g}} \right\|} \right.}}{{2\sigma _g^2}}} \right],j = 1,2,...,s,g = 1,2,...,{m_{\rm{2}}}$;
            \State $y_{ij}^{} = \sum\limits_{k = 1}^{{m_{\rm{1}}}} {w_{kj}^{(2)}} ne{t_{ikj}} + b_j^{(2)},j = 1,2,...,s$;
            \State $r_{ij}^{} = \sum\limits_{g = {\rm{1}}}^{{m_{\rm{2}}}} {w_{gj}^{{\rm{(2)}}}} ne{t_{igj}} + b_j^{(2)},j = 1,2,...,s$;
            \State Minimization objective function equation and updates the model weights $w_k$ and threshold $b_k$;
            \If{$x_{ij}$ is the sample containing the missing values} 
              \State Update the missing value variables in $x_{ij}$ based on Adam optimization algorithm. The missing value variables will be dynamically optimised in collaboration with the model parameters;
            \EndIf
            \State Continue to take the next sample in $X_b$;
          \Until{After traversing all samples in $X_b$}
          \State  The final output of the algorithm is the complete dataset $Y_{ij}$, where the model output $y_{ij}$ corresponding to the missing values is used as the final fill value.
         \Until{Termination of training conditions}
         
\end{algorithmic}
\end{algorithm} 
FFEAM first fills the missing data set with random forest in advance. Then, the MVDC fill strategy treats the entire incomplete dataset as the training set and includes the missing values as variables in the loss function, while the $h$ centroids $\mu_h$ are selected according to the k-means algorithm, and the width $\sigma_g$ is calculated by Equation (6). 
\par As we treat missing values as variables and include incomplete samples and complete samples in network training, the missing value variables and model parameters are dynamically optimized together. The advantage of the MVDC filling scheme lies in its dynamic optimization of missing values, allowing the network model to gradually match the regression structure in incomplete data. With the increase of optimization depth, the filling accuracy of missing values and the accuracy of the model will also be improved.
\par The samples that contain dynamic fill values are then fed into FFEAM, which carries out two-dimensional fusion learning of the interrelated features and common features in the data. The pseudo code is shown in Algorithm 1.

\section{Experiments}\label{s5.0}

In this section, we first conduct experiments on four UCI (University of California Irvine) datasets to evaluate the performance of the proposed method. Secondly, we perform extensive additional experiments on three more complex real datasets to analyze the effectiveness of our model, and the effectiveness of the model was verified by comparing the filling performance with the mean filling method, Autoencoder \cite{4}, KNN \cite{22}, MIDAS \cite{23} and CE-AANN \cite{8}. We also constructed six artificial datasets and conducted six model comparison experiments on these to analyse the effect of information such as noise and data dimensionality on the models. Finally, we also performed time complexity analysis on the FFEAM model.
\par Firstly, various types of deep learning models were constructed based on Tensorflow, including the proposed FFEAM, and the Scikit-learn machine learning library was used to construct the mean-fill model, and all experiments were conducted on a PC computer with an operating system of Windows 10 Professional, a processor of Intel(R) Core(TM) i7-7700HQ CPU @2.80 GHz quad-core, 16 GB of RAM (DDR4 2400 MHz), and SD8SN8U- 128G -1006 (128 GB) as the main hard disk.

\subsection{Datasets}
\par In this paper, we have conducted relevant experiments in 13 datasets. For each dataset we set a different missing rate, the whole incomplete dataset was used as the training set and the missing values were used as variables for the loss function.
\par The first four datasets used in the experiments are open-source datasets originating from UCI, as detailed in Table \ref{tbl:rescomp0}, and the experiments were conducted by randomly removing some existing values from the complete data, thus constructing incomplete datasets with missing rates set to 20\%, 30\%, 40\%, and 50\%, respectively.

\begin{table}[htbp]\centering
\caption{Description of the UCI experimental dataset}
\label{tbl:rescomp0}
\begin{tabular}{@{}lll@{}}
\toprule
Data set name & Sample size & Number of attributes \\ 
\midrule
Iris  & 150  & 4    \\
Wine  & 178  & 14   \\
Cloud & 1024 & 10   \\
Seeds & 210  & 7    \\ 
\bottomrule  
\end{tabular}
\footnotetext{Source: The data in the above table are all from the public UCI dataset: $https://archive.ics.uci.edu/ml/index.php$}
\end{table}

\begin{table}\centering
\caption{Description of complex experimental dataset}
\label{tbl:rescomp1}
\begin{tabular}{@{}lll@{}}
\toprule
Data set name                                     & Sample size & Number of attributes \\ \midrule
Traffic data of Baoan District\footnotemark[1]                  & 1102        & 10                   \\
Beijing PM2.5Data Data Set\footnotemark[2]                        & 43824       & 13                   \\
AI4I 2020 Predictive Maintenance Dataset Data Set\footnotemark[3] & 10000       & 14                  \\ \bottomrule

\end{tabular}
\footnotetext[1]{The data comes from traffic flow data in Baoan district released by the Shenzhen municipal government's open platform: 
$https://opendata.sz.gov.cn/data/dataSet/toDataDetails/29200_02803199$}
\footnotetext[2]{The data comes from Beijing's air quality data set:
$http://www.bjmemc.com.cn/$}
\footnotetext[3]{This data comes from the AI4I 2020 Predictive Maintenance dataset:
$https://archive.ics.uci.edu/ml/datasets/AI4I+2020+Predictive+Maintenance+Dataset\#$}
\end{table}

Secondly, extensive experiments were also conducted on three more complex real datasets (missing rate set to 20\%), as detailed in Table \ref{tbl:rescomp1}. the first dataset is the traﬀic flow data of Baoan district released by the Shenzhen government open platform in China, which spans from November 2017 to May 2021, and contains 10 attributes with a total of 1102 samples. The second one is the Beijing PM2.5Data Data Set, which contains air quality data of Beijing from 2010-2014, with a time interval of 1 hour, 13 attributes, and 43824 samples. The third dataset is the AI4I 2020 Predictive Maintenance Dataset, which is a comprehensive dataset reflecting real predictive maintenance data encountered in the industry, and contains 14 attributes with 10,000 samples.

\begin{figure}[H]
\center{\includegraphics[width=\textwidth] {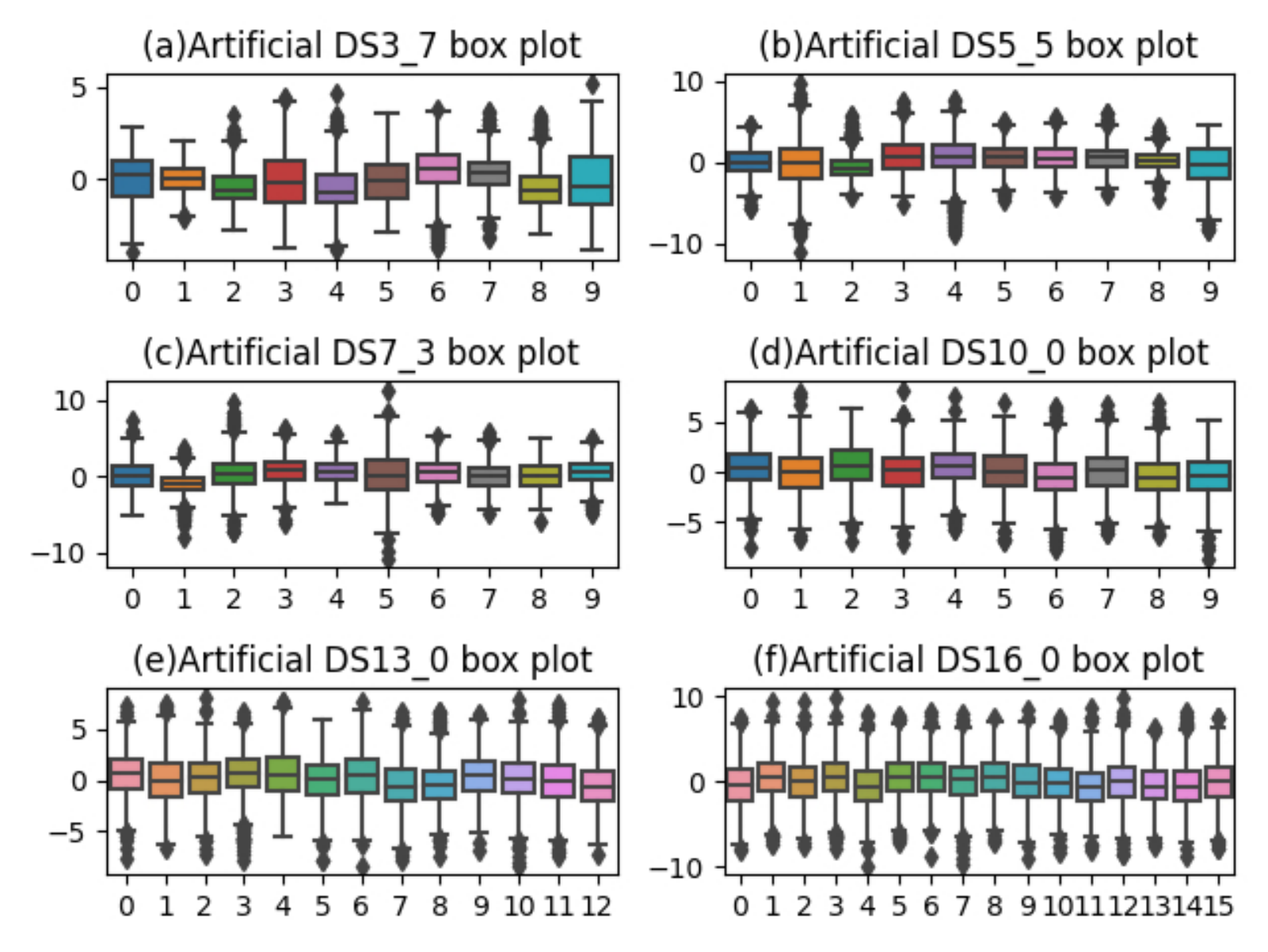}}
\caption{Box plot of artificial data}
\label{ref8}
\end{figure}

Finally, we manually constructed six artificial datasets to validate the effect of noise, data dimensionality on the effect of the model. All six datasets consisted of 1000 samples, and their box plots are shown in Figure \ref{ref8}. Figure (a) represents the artificial data DS3\_7,with a total of 10 features, of which 3 features are valid and 7 features are random noise features. Figure (b) shows the artificial data DS5\_5, with a total of 10 features, of which 5 are valid and 5 are random noise features. Figure (c) shows the artificial data DS7\_3, with a total of 10 features, of which 7 are valid and 3 are random noise features. Figure (d) shows the artificial data DS10\_0, with a total of 10 features, of which all 10 features are valid. Figure (e) shows the artificial data DS13\_0, with 13 features in total, of which all features are valid. Figure (f) shows the artificial data DS16\_0, with 16 features in total, and all features are valid.

\subsection{Baseline model and hyperparameter settings}

 To validate the filling performance of the model, FFEAM was compared with five benchmark models of mean filling (Means),  Autoencoder\cite{4},KNN\cite{22}, MIDAS\cite{23}, and CE-AANN\cite{8}.
\par Means: a classical traditional statistical filling method where numerical data is filled with the average of all existing values in the incomplete attribute column.
\par AE\cite{4}:Based on classical autoencoder for missing value filling, autoencoder are generally used for feature learning where we fill in the missing values by reproducing the input values at the output.
\par KNN\cite{22}:KNN finds the K complete samples that are closest to or most correlated with each incomplete sample and uses the weighted average of the existing values of each complete sample as the fill value.
\par MIDAS\cite{23}:An accurate, fast, and scalable approach to multiple imputation, which call MIDAS (Multiple Imputation with Denoising Autoencoders) proposed by Lall et al. 
\par CE-AANN\cite{8}: A correlation-enhanced self-encoder filling model proposed by Lai et al.
\par To ensure the fairness of the experiments, the base hyperparameters of the benchmark comparison model and the FFEAM model were the same for each dataset with different missing rates, specifically set as follows: learning rate of 0.1, number of training iterations of 1000, batch size of 20, and the total number of hidden layer neurons of 20. Since in the correlation-enhanced self-encoder model, two types of hidden layer neurons are contained, the specific hidden layer The neuron assignments are both $m_1$=10 and $m_2$=10. Adam function is used as the training optimizer for all models.

\subsection{Evaluation Indicators}
\label{}
Root Mean Square Error (RMSE) and Mean Absolute Error (MAE) are used as evaluation indexes for filling performance, and the formulas for RMSE and MAE are Equation (12) and (13), respectively.

\begin{equation}
RMSE = \sqrt {\frac{{\rm{1}}}{n}\sum\limits_{i = 1}^n {\mathop {({x_i} - {y_i})}\nolimits^2 } },
\label{pythagorean}
\end{equation}

\begin{equation}
MAE = \frac{{\rm{1}}}{n}\sum\limits_{i = {\rm{1}}}^n {\mathop {\left\| {\left. {{x_i} - {y_i}} \right\|} \right.}\nolimits^{} },
\label{pythagorean}
\end{equation}
in Equation (12) and (13), $n$ is the total number of samples, $y_i$ denotes the filled value, and $x_i$ denotes the true value corresponding to that filled value.

\subsection{Analysis of experimental results} 
\label{}

Experiments were first conducted based on four UCI datasets, and Table \ref{tbl:rescomp2} depicts the filling errors of different models for different datasets with different missing rates. From Table \ref{tbl:rescomp2}, it can be seen that FFEAM outperforms the other comparison models in terms of missing value filling. According to the experimental results, the KNN-based filling method is better than the traditional mean-value filling and AE methods, mainly because it selects the K samples that are closest to the missing values and fills the missing values based on these K samples, making better use of the common features among the samples. However, compared with KNN, FFEAM achieves the lowest RMSE and MAE for different missing rates, for example, in the Wine dataset with 20\% missing rate, the RMSE and MAE of FFEAM are 2.73 and 0.261 lower than KNN respectively. Based on the comparison of the experimental results of FFEAM and CE-AANN, it can be seen that the filling accuracy of FFEAM is improved in all four datasets with different missing rates. For example, when there is a 20\% missing rate, FFEAM can reduce the RMSE values by 0.032, 2.984, and 0.102 compared to CE-AANN on the Iris, Wine, Cloud, and Seeds datasets, respectively, while MAE decreased by 0.008, 0.222, 0.531, and 0.007, respectively. Combining the performance of the other datasets, the results of FFEAM in most cases, achieved the lowest RMSE and MAE still outperformed MIDAS. This is mainly due to the fact that the MIDAS method mainly fills missing values from the data interrelated feature dimension without considering the common features, and FFEAM combines both types of features to fill missing values, thus improving the filling accuracy.
\begin{table}\centering
\caption{Comparison of filling errors of different models}
\label{tbl:rescomp2}
\resizebox{\textwidth}{50mm}{
\begin{tabular}{@{}llllllllll@{}}
\cmidrule(r){1-10}
\multirow{2}{*}{Data set name} & \multirow{2}{*}{Model name} & \multicolumn{4}{l}{RMSE}          & \multicolumn{4}{l}{MAE}           \\ \cmidrule(lr){3-10}
                       &                       & 20\%   & 30\%   & 40\%   & 50\%   & 20\%   & 30\%   & 40\%   & 50\%   \\ \cmidrule(r){1-10}
\multirow{6}{*}{Iris}  & MEANS                 & 0.465  & 0.517  & 0.633  & 0.674  & 0.149  & 0.205  & 0.275  & 0.328  \\
                       & AE                    & 0.277  & 0.355  & 0.495  & 0.407  & 0.083  & 0.126  & 0.205  & 0.199  \\
                       & KNN                   & 0.228  & 0.274  & 0.386  & 0.351  & 0.063  & 0.098  & 0.147  & 0.159  \\ 
                       & MIDAS                   & 0.151  & 0.213  & 0.339  & 0.343  & 0.048  & 0.083  & 0.136  & 0.151  \\ 
                       & CE-AANN               & 0.161  & 0.287  & 0.435  & 0.355  & 0.049  & 0.089  & 0.145  & 0.144  \\ 
                       & FFEAM                & $\bm{0.129}$  & $\bm{0.175}$  & $\bm{0.335}$  & $\bm{0.294}$  & $\bm{0.041}$ & $\bm{0.064}$ & $\bm{0.126}$ & $\bm{0.129}$ \\ 
\cmidrule(lr){1-10}
\multirow{6}{*}{Wine}  & MEANS                 & 32.616 & 42.534 & 52.027 & 48.024 & 3.405  & 5.035  & 6.931  & 7.420  \\
                       & AE                    & 32.669 & 42.502 & 51.894 & 47.605 & 3.432  & 5.056  & 7.007  & 7.469  \\
                       & KNN                   & 25.510 & 27.932 & 35.179 & 36.678  & 2.644  & 3.423  & 4.537 & 5.623  \\
                       & MIDAS                   & 25.487  & 26.539  & 37.571  & 38.946  & 2.732  & 3.152  & 4.579  & 5.637  \\ 
                       & CE-AANN               & 25.764 & 35.115 & 44.731 & 44.193 & 2.605  & 4.084  & 5.669  & 6.547  \\
                       & FFEAM                  & $\bm{22.780}$ & $\bm{24.553}$ & $\bm{29.518}$ & $\bm{35.272}$ & $\bm{2.383}$ & $\bm{3.226}$ & $\bm{4.044}$ & $\bm{5.456}$ \\
\cmidrule(lr){1-10}
\multirow{6}{*}{Cloud} & MEANS                 & 60.972 & 69.023 & 86.299 & 87.653 & 13.607 & 19.073 & 25.100 & 29.261 \\
                       & AE                    & 59.680 & 53.559 & 72.482 & 72.233 & 6.804  & 7.858  & 15.558 & 15.345 \\
                       & KNN                   & 29.084 & 38.730 & 58.432 & 55.562  & 3.573  & 8.079 & 13.901 & 15.960  \\
                       & MIDAS                   & 26.549 & 37.481 & 45.609 & 50.959  & 3.596  & 5.992 & 8.463 & 11.396  \\
                       & CE-AANN               & 17.148 & 29.276 & $\bm{28.669}$ & 37.128 & 3.473  & 5.388  & 6.504  & 10.627 \\
                       & FFEAM                  & $\bm{15.516}$ & $\bm{28.777}$ & 28.944  & $\bm{32.532}$ & $\bm{2.942}$ & $\bm{5.225}$         & $\bm{5.920}$ & $\bm{8.655}$ \\
\cmidrule(lr){1-10}
\multirow{6}{*}{Seeds} & MEANS                 & 0.600  & 0.697  & 0.782  & 0.822  & 0.155  & 0.220  & 0.292  & 0.332  \\
                       & AE                    & 0.606  & 0.719  & 0.794  & 0.821  & 0.158  & 0.234  & 0.299  & 0.339  \\
                       & KNN       & 0.322 & 0.488 & 0.442 & 0.503 & 0.064  & 0.130 & 0.146 & 0.183  \\
                       & MIDAS       & 0.264 & 0.350 & 0.381 & 0.492 & $\bm{0.061}$  & $\bm{0.095}$ & $\bm{0.119}$ & 0.173  \\
                       & CE-AANN               & 0.363  & 0.564  & 0.444  & 0.589  & 0.073  & 0.141  & 0.147  & 0.199  \\
                       & FFEAM                 & $\bm{0.261}$  & $\bm{0.336}$  & $\bm{0.379}$  & $\bm{0.482}$  & 0.066 & 0.109 & 0.127 & $\bm{0.179}$ \\ \bottomrule
\end{tabular}}
\end{table}

\par
\begin{figure}[H]
\center{\includegraphics[width=1\textwidth] {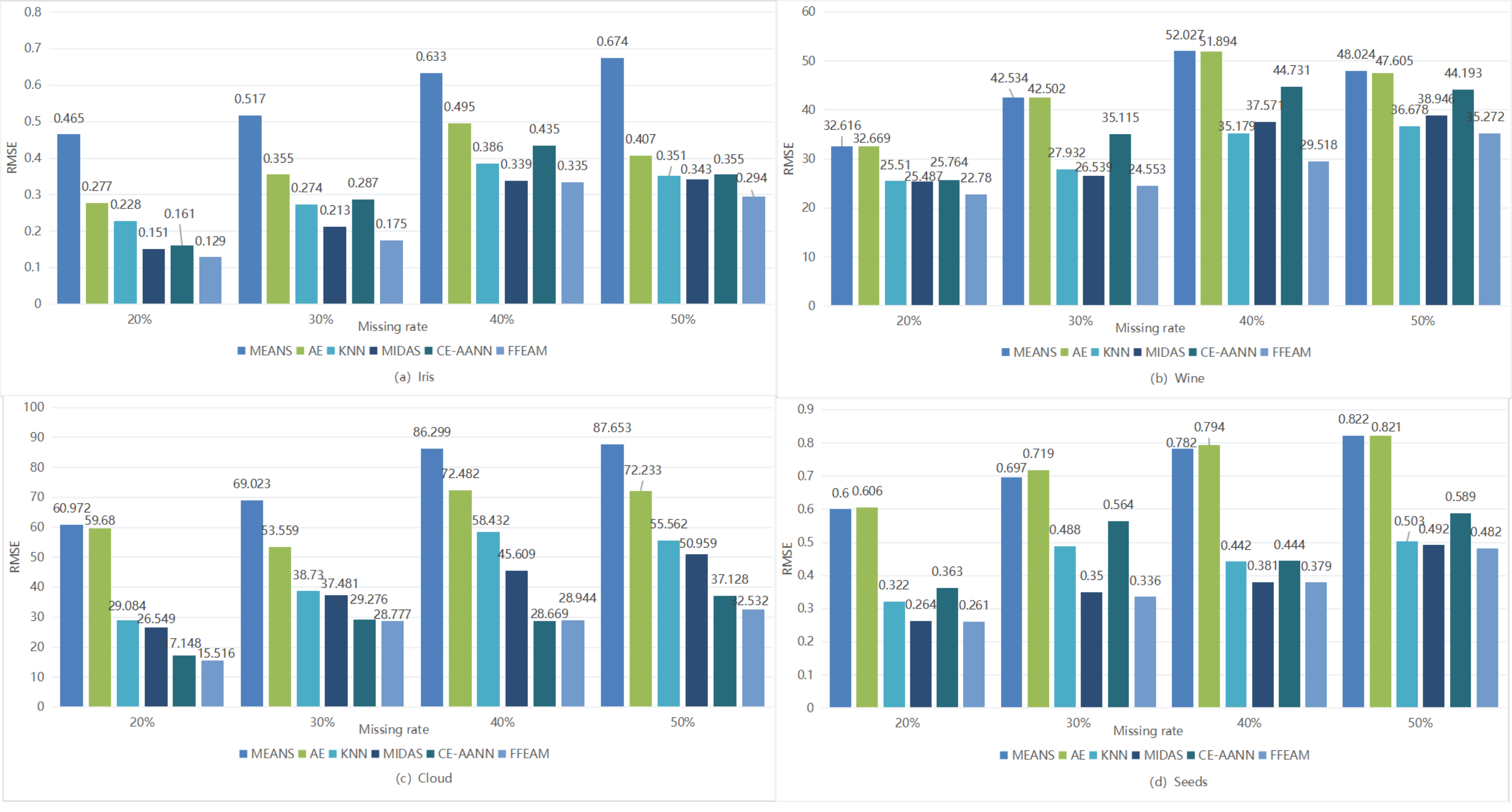}}
\caption{Comparison of RMSE of different models on four UCI datasets}
\label{ref3}
\end{figure}

\begin{figure}[H]
\center{\includegraphics[width=1\textwidth] {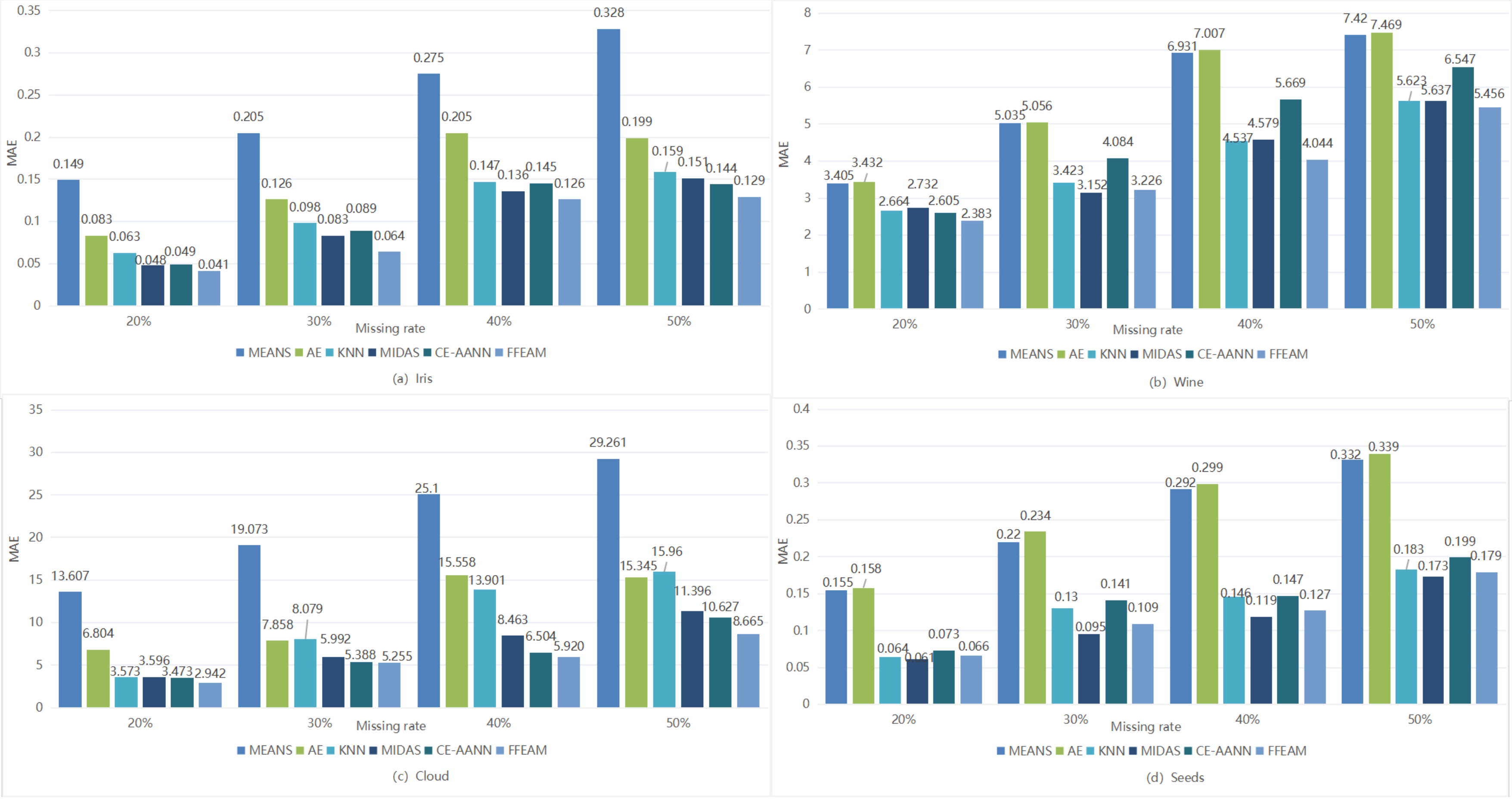}}
\caption{Comparison of MAE of different models on four UCI datasets}
\label{ref4}
\end{figure}

 Figure \ref{ref3} shows the differences in RMSE metrics of the six models on the different datasets, and Figure \ref{ref4} shows the differences in MAE metrics. According to  Figure \ref{ref3} and  Figure \ref{ref4},  it can be seen that the error metrics of  AE are close to Means in Wine and Seed datasets, while in the other two datasets, the error of AE decreases compared to Means, indicating that the filling effect of AE is better than Means, and it also shows that AE itself has self-tracking and lacks the ability to discover the interrelated and common features among attributes. In addition, the KNN method is superior to the Means and AE methods, mainly because of the KNN's ability to discover the common feature dimensions of the data. It can also be seen from the figure that the vast majority of sub-optimal results come from MIDAS, with CE-AANN being slightly inferior to MIDAS, while the best performer is FFEAM. This is mainly because the model has a better fusion learning ability of data interrelated features and common features, which further improves the performance of the missing values filling model.

\par In addition, three more complex real dataset comparison experiments were conducted, based on the traﬀic flow data of Baoan District, Beijing PM2.5Data Data Set and AI4I 2020 Predictive Maintenance Dataset Data Set, and the incomplete dataset with 20\% missing rate was randomly constructed and compared with the five benchmark models. The experimental results obtained are shown in Table \ref{tbl:rescomp4}. We can observe that FFEAM outperforms the benchmark models in terms of filling missing values on the real dataset with the lowest RMSE and MAE. In the Beijing PM2.5Data Data Set experiment, the RMSE of FFEAM decreased by 1.639, 1.660, 4.157, 2.579, and 0.686 compared to MEANS, AE, KNN, MIDAS, and CE-AANN, respectively. In the AI4I 2020 Predictive Maintenance Dataset Data Set, the RMSE of FFEAM was reduced by 9.29, 9.288, 9.116, 3.832, and 6.207 compared to MEANS, AE, KNN, MIDAS, and CE-AANN, respectively;

\begin{table}\centering
\caption{Comparison of filling errors of different models in real data sets}
\label{tbl:rescomp4}
{\resizebox{\linewidth}{!}{
\begin{tabular}{@{}cccc@{}}
\toprule
Data set name                                                      & Model Name & RMSE             & MAE             \\ \midrule
\multirow{6}{*}{Traffic data of Baoan District}                    & MEANS      & 1687.9270        & 286.815         \\
                                                                   & AE         & 1694.582         & 278.533         \\
                                                                   & KNN        & 875.594          & 115.499         \\
                                                                    & MIDAS        & 916.408          & 127.106         \\
                                                                   & CE-AANN    & 807.186          & 101.557          \\
                                                                   & FFEAM      & $\bm{593.246}$ & $\bm{85.458}$ \\
\cmidrule(lr){1-4}
\multirow{6}{*}{Beijing PM2.5Data Data Set}                        & MEANS      & 13.746           & 1.541           \\
                                                                   & AE         & 13.767           & 1.568           \\
                                                                   & KNN        & 16.264           & 2.958         \\
                                                                   & MIDAS        & 14.686           & 1.463         \\
                                                                   & CE-AANN    & 12.793           & 1.484           \\
                                                                   & FFEAM      & $\bm{12.107}$  & $\bm{1.321}$  \\
\cmidrule(lr){1-4}
\multirow{6}{*}{AI4I 2020 Predictive Maintenance Dataset Data Set} & MEANS      & 24.716           & 3.183           \\
                                                                   & AE         & 24.714           & 3.195           \\
                                                                   & KNN        & 24.542           & 3.074         \\
                                                                   & MIDAS       & 19.258           & 3.191         \\
                                                                   & CE-AANN    & 21.633           & 3.153           \\
                                                                   & FFEAM      & $\bm{15.426}$  & $\bm{2.601}$  \\ \bottomrule

\end{tabular}}}
\end{table}

\par
\begin{figure}[H]
\center{\includegraphics[width=\textwidth] {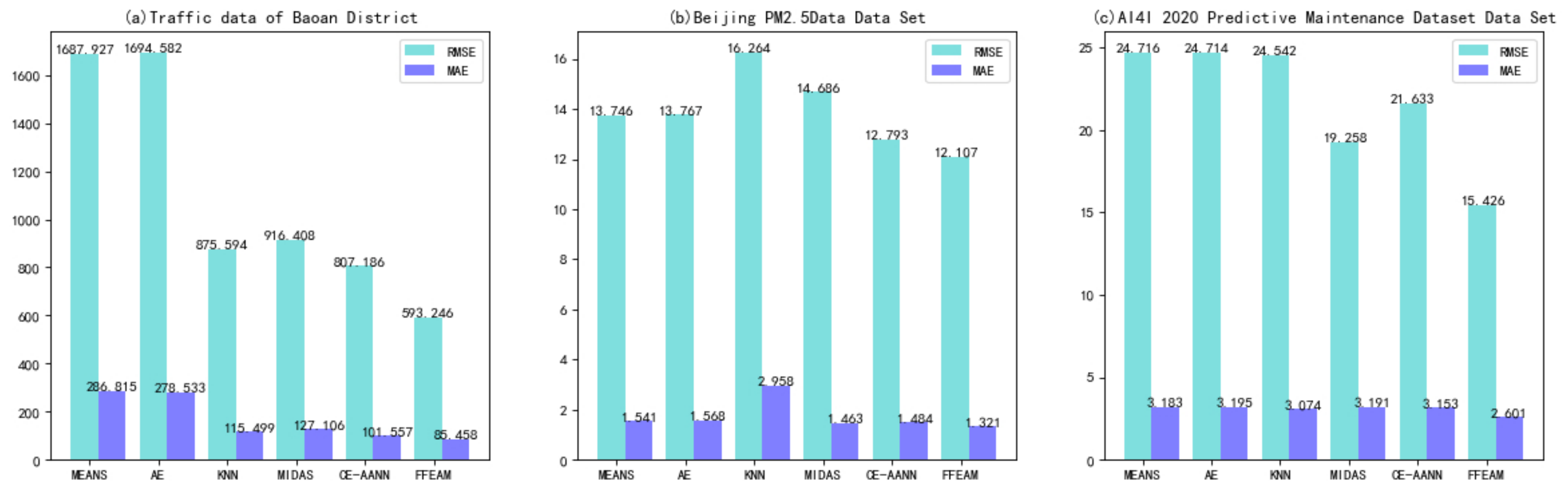}}
\caption{Comparison histogram of filling errors of different models in real data sets}
\label{ref5}
\end{figure}

 Figure \ref{ref5} shows the differences in RMSE and MAE metrics for the six models on three complex real data sets. The visual comparison analysis in the figure shows that MEANS and AE perform comparable results, and their RMSE and MAE are not very different, KNN and MIDAS have a reduced effect on larger data sets, while the CE-AANN model maintains good performance with large amounts of data, with lower RMSE and MAE than KNN and MIDAS, and greater robustness. However, the best results on these three real datasets still come from the model FFEAM.

\begin{table}\centering
\caption{Comparison of filling errors of different models in Artificial datasets}
\label{tbl:rescomp8}
{\setlength{\tabcolsep}{9mm}{
\begin{tabular}{@{}cccc@{}}
\toprule
Data set name                                                      & Model Name & RMSE             & MAE             \\ \midrule
\multirow{6}{*}{Artificial DS3\_7}                    & MEANS      & 0.452        & 0.153         \\
                                                                   & AE         &  0.546         & 0.185         \\
                                                                   & KNN        & 0.439         &  0.148         \\
                                                                    & MIDAS        & 0.459          & 0.154         \\
                                                                   & CE-AANN    & 0.458          & 0.152          \\
                                                                   & FFEAM      & $\bm{0.437}$ & $\bm{ 0.147}$ \\
\cmidrule(lr){1-4}
\multirow{6}{*}{Artificial DS5\_5}                        & MEANS      &  0.596           & 0.198           \\
                                                                   & AE         & 0.682           & 0.227           \\
                                                                   & KNN        & 0.565           & 0.187         \\
                                                                   & MIDAS        &  0.567           & 0.189        \\
                                                                   & CE-AANN    & 0.571           & 0.189           \\
                                                                   & FFEAM      & $\bm{0.562}$  & $\bm{0.187}$  \\
\cmidrule(lr){1-4}
\multirow{6}{*}{Artificial DS7\_3}                       & MEANS      & 0.638           &  0.212           \\
                                                                   & AE         & 0.758           & 0.254           \\
                                                                   & KNN        & 0.591          & 0.196         \\
                                                                   & MIDAS        & 0.594           & 0.197         \\
                                                                   & CE-AANN    & 0.596           & 0.198           \\
                                                                   & FFEAM      & $\bm{0.576}$  & $\bm{0.190}$  \\ 
\cmidrule(lr){1-4}
\multirow{6}{*}{Artificial DS10\_0}                       & MEANS      & 0.881           &  0.297           \\
                                                                   & AE         & 1.038           & 0.349           \\
                                                                   & KNN        &  0.836          &  0.282         \\
                                                                   & MIDAS        & 0.806           &  0.270         \\
                                                                   & CE-AANN    & 0.891           & 0.297           \\
                                                                   & FFEAM      & $\bm{0.769}$  & $\bm{ 0.255}$  \\ 
\cmidrule(lr){1-4}
\multirow{6}{*}{Artificial DS13\_0}                       & MEANS      & 0.951           & 0.321           \\
                                                                   & AE         & 1.182           & 0.393           \\
                                                                   & KNN        &  0.902          &  0.301         \\
                                                                   & MIDAS        & 0.881         &   0.293       \\
                                                                   & CE-AANN    &  0.923           & 0.305           \\
                                                                   & FFEAM      & $\bm{0.826}$  & $\bm{ 0.273}$  \\ 
\cmidrule(lr){1-4}                                                               
\multirow{6}{*}{Artificial DS16\_0}                       & MEANS      & 1.060           & 0.357           \\
                                                                   & AE         & 1.355          & 0.458           \\
                                                                   & KNN        & 0.997          &  0.335       \\
                                                                   & MIDAS        & 0.955         &  0.313        \\
                                                                   & CE-AANN    &  1.051           & 0.352           \\
                                                                   & FFEAM      & $\bm{0.935}$  & $\bm{  0.313}$  \\ 
                                                                   \bottomrule

\end{tabular}}}
\end{table}

\par We also tested six models on six artificial datasets, where we set the missing rate to 20\% for each dataset, and the results are shown in the Table \ref{tbl:rescomp8}. In the experiments on datasets DS3\_7, DS5\_5, DS7\_3, FFEAM achieved the lowest RMSE and MAE for all three datasets, which consisted of 70\%, 50\% and 30\% noisy vectors respectively, indicating that FFEAM can maintain high filling accuracy even in data containing noise. Similarly, in the experiments on datasets DS10\_0, DS13\_0, DS16\_0, FFEAM outperformed other methods by achieving the lowest RMSE and MAE values. The experiments proved that FFEAM still achieved better filling results as the feature dimension was increased. In the aforementioned experiment, the presence of noise vectors had a significant impact on the learning and extraction of interrelated features between attributes. therefore, methods such as CE-AANN and MIDAS that fill in interrelated features from attribute dimensions did not perform as well as KNN. As feature dimensions increased, common features between data became more complex, and therefore filling KNN with common feature dimensions led to a decrease in performance. FFEAM maintains the best filling results in both of these cases, mainly because it learns from the advantages of both dimensions simultaneously and performs imputation by incorporating both common and interrelated features.

\begin{table}\centering
\caption{Statistical test of different models on artificial data sets}
\label{tbl:rescomp16}
{\setlength{\tabcolsep}{9mm}{
\begin{tabular}{ccc}
\hline
Data sets                           & Model Name & P              \\ \hline
\multirow{6}{*}{Artificial DS3\_7}  & MEANS      & 0.954          \\
                                    & AE         & 0.731          \\
                                    & KNN        & 0.981          \\
                                    & MIDAS      & 0.944          \\
                                    & CE-AANN    & 0.952          \\
                                    & FFEAM      & \textbf{0.987} \\ \hline
\multirow{6}{*}{Artificial DS5\_5}  & MEANS      & 0.809          \\
                                    & AE         & 0.631          \\
                                    & KNN        & 0.929          \\
                                    & MIDAS      & 0.908          \\
                                    & CE-AANN    & 0.854          \\
                                    & FFEAM      & \textbf{0.978} \\ \hline
\multirow{6}{*}{Artificial DS7\_3}  & MEANS      & 0.810          \\
                                    & AE         & 0.729          \\
                                    & KNN        & 0.969          \\
                                    & MIDAS      & 0.911          \\
                                    & CE-AANN    & 0.875          \\
                                    & FFEAM      & \textbf{0.982} \\ \hline
\multirow{6}{*}{Artificial DS10\_0} & MEANS      & 0.761          \\
                                    & AE         & 0.742          \\
                                    & KNN        & 0.972          \\
                                    & MIDAS      & 0.916          \\
                                    & CE-AANN    & 0.772          \\
                                    & FFEAM      & \textbf{0.987} \\ \hline
\multirow{6}{*}{Artificial DS13\_0} & MEANS      & 0.818          \\
                                    & AE         & 0.781          \\
                                    & KNN        & 0.953          \\
                                    & MIDAS      & 0.941          \\
                                    & CE-AANN    & 0.903          \\
                                    & FFEAM      & \textbf{0.976} \\ \hline
\multirow{6}{*}{Artificial DS16\_0} & MEANS      & 0.865          \\
                                    & AE         & 0.749          \\
                                    & KNN        & 0.926          \\
                                    & MIDAS      & 0.925          \\
                                    & CE-AANN    & 0.873          \\
                                    & FFEAM      & \textbf{0.968} \\ \hline
\end{tabular}}}
\end{table}

In addition, we carried out statistical testing on six artificial data sets utilizing the T-Test from STAC \cite{24}. We conducted a T-Test separately between the original data (complete data set) and the data filled in by each model. The outcomes of this test are displayed in Table \ref{tbl:rescomp16}. A higher p-value observed during the T-Test indicates a smaller difference between the model filled data and the original data. This means that there is no significant disparity between them, and FFEAM yielded the highest P-value across all baseline algorithms, implying that the data filled by FFEAM differed the least from the original data.
\par To investigate the effect of the specific assignment of $m_1$ and $m_2$ values on the filling effect of the FFEAM model when the total number of neurons in the hidden layer is constant, experiments were conducted on the Iris dataset with different deletion rates. By setting the total number of neurons constant and fixed to 20, i.e., $m_1$+$m_2$=20, the experimental error results were obtained by changing the ratio of $m_1$ and $m_2$, as shown in Table \ref{tbl:rescomp5}. It can be seen through Table \ref{tbl:rescomp5} that the optimal results with different deletion rates are mostly clustered around $m_1$=9 and $m_2$=11, and there is not much fluctuation. And combined with Table \ref{tbl:rescomp2}, it can be seen that the worst performance of the FFEAM model with different missing rates is also better than the second-best result of CE-AANN with the same missing rate in Table \ref{tbl:rescomp2}, for example, in the case of 20\% missing rate, the highest RMSE of the FFEAM model is 0.145 and the highest MAE is 0.045, while in CE-AANN, the RMSE is 0.161 and the MAE is 0.049, once again validating the filling performance of the FFEAM model.

\begin{table}[!h]\centering
\caption{Comparison of filling errors for different $m_1$ and $m_2$ assignments}
\label{tbl:rescomp5}
\begin{tabular}{@{}lllllllll@{}}
\toprule
\multirow{2}{*}{$m_1$,$m_2$} & \multicolumn{4}{c}{RMSE}      & \multicolumn{4}{c}{MAE}       \\ \cmidrule(l){2-9} 
                       & 20\%  & 30\%  & 40\%  & 50\%  & 20\%  & 30\%  & 40\%  & 50\%  \\ \cmidrule(r){1-9}
5,15                   & 0.205 & 0.218 & 0.425 & 0.385 & 0.058 & 0.085 & 0.147 & 0.149 \\
6,14                   & 0.145 & 0.188 & 0.422 & 0.340 & 0.045 & 0.073 & 0.138 & 0.141 \\
7,13                   & 0.139 & 0.183 & 0.379 & 0.348 & 0.043 & 0.071 & 0.123 & 0.145 \\
8,12                   & 0.133 & $\bm{0.168}$ & 0.369 & 0.344 & 0.042 & 0.067 & 0.30  & 0.137 \\
9,11                   & $\bm{0.128}$ & 0.171 & $\bm{0.320}$ & 0.311 & $\bm{0.040}$ & 0.068 & $\bm{0.117}$ & 0.131 \\
10,10                  & 0.129 & 0.175 & 0.335 & $\bm{0.294}$ & 0.041 & $\bm{0.064}$ & 0.126 & $\bm{0.129}$ \\
11,9                   & 0.131 & 0.176 & 0.358 & 0.332 & 0.042 & 0.069 & 0.124 & 0.142 \\
12,8                   & 0.137 & 0.177 & 0.393 & 0.333 & 0.043 & 0.070 & 0.128 & 0.138 \\
13,7                   & 0.139 & 0.175 & 0.387 & 0.357 & 0.046 & 0.068 & 0.139 & 0.155 \\
14,6                   & 0.140 & 0.180 & 0.399 & 0.387 & 0.044 & 0.070 & 0.133 & 0.158 \\
15,5                   & 0.141 & 0.182 & 0.407 & 0.402 & 0.045 & 0.070 & 0.127 & 0.154 \\ \bottomrule
\end{tabular}
\end{table}

\par
\begin{figure}[H]
\center{\includegraphics[width=0.70\textwidth] {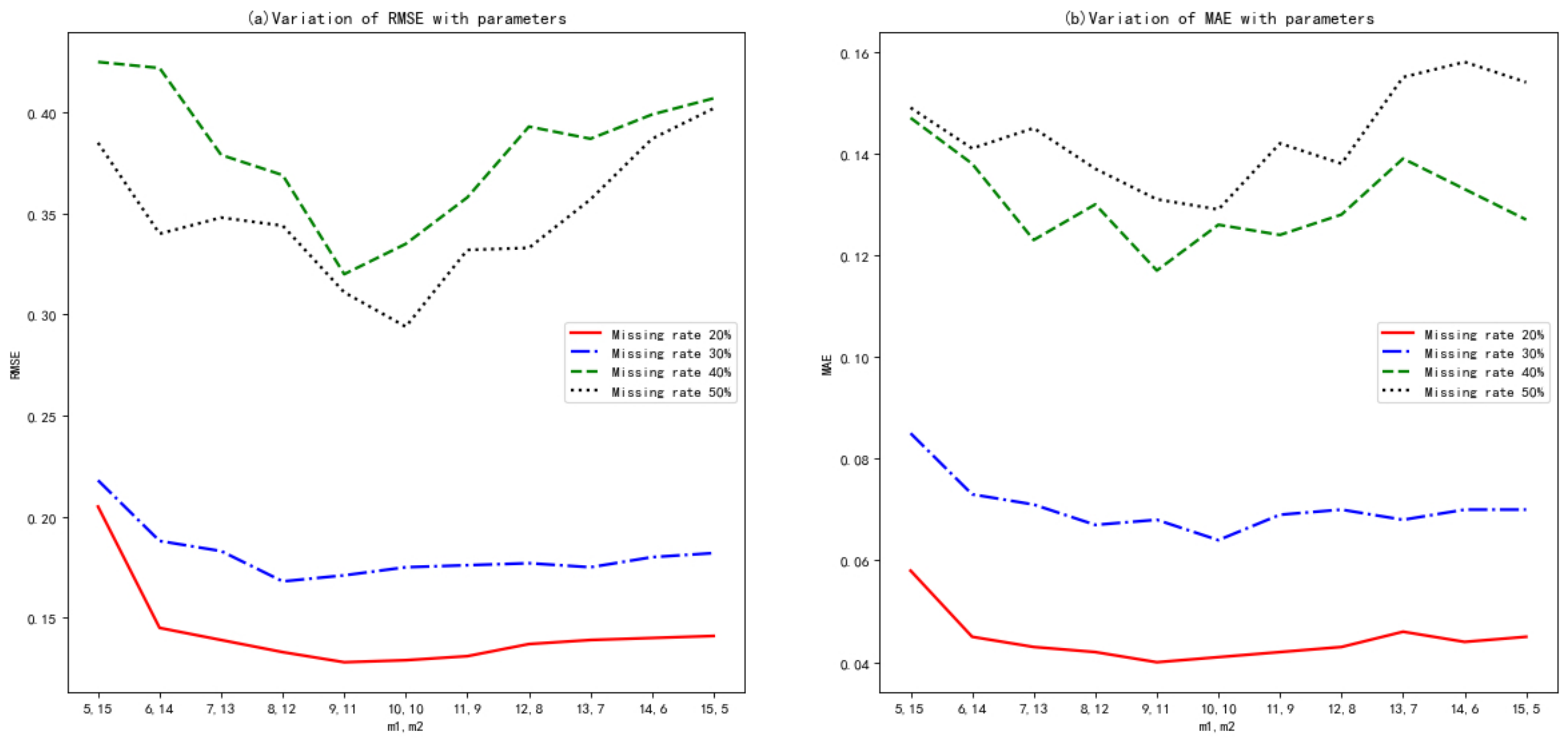}}
\caption{Comparison of filling errors for different $m_1$ and $m_2$ assignments}
\label{ref6}
\end{figure}

  Figure \ref{ref6} shows the line graph of filling error comparison for different $m_1$ and $m_2$ assignments. Through Figure \ref{ref6}, it can be visualized that the model filling error, with the total number of hidden layer neurons unchanged, first shows a decreasing trend as $m_1$ increases and $m_2$ decreases, and then shows an increasing trend when it decreases to a certain value. The main reason is that when the value of $m_1$ is too small, the number of de-tracking neurons is too small to effectively explore the interrelated features among data attributes, so as the number of $m_1$ increases, the ability to explore the association features is improved and the model error is reduced. And when the value of $m_2$ is too small, the number of radial basis function neurons is too small, and the ability of the model to mine the common features of the data is insufficient, which leads to an increase in the model error. Therefore, the proposed mutual design of the two types of neuron hidden layers needs to set the values of $m_1$, $m_2$ reasonably in order to support the cooperative fusion learning of data common features and interrelated features. 

\par In summary, the proposed FFEAM has better filling performance than the benchmark models regardless of the settings for different missing rates or on different datasets. The main reason lies in the fact that the interaction design between de-tracking neurons and radial basis function neurons allows the model to perform fusion learning of data interrelated features and data common features simultaneously.

\subsection{Time complexity analysis} 
We performed time complexity experiments on three datasets: Traffic data of Baoan District, AI4I 2020 Predictive Maintenance Dataset, and Cloud. For all three datasets, we kept the hyperparameters of FFEAM, CE-AANN, and AE identical, and the results are presented in Table \ref{tbl:rescomp6}.

\begin{table}[!h]\centering
\caption{Performance comparison of FFEAM, CE-AANN and AE on various datasets in terms of computing time and MAE}
\label{tbl:rescomp6}
\resizebox{\textwidth}{12mm}{
\begin{tabular}{ccccccc}
\hline
\multirow{2}{*}{Data sets}                        & \multicolumn{3}{c}{Times(seconds)} & \multicolumn{3}{c}{MAE}     \\ \cline{2-7} 
                                                  & FFEAM      & CE-AANN   & AE        & FFEAM   & CE-AANN & AE       \\ \hline
Traffic data of Baoan District                    & 91.0942    & 90.8161   & 65.3581   & 85.458 & 101.557 & 278.533 \\
AI4I 2020 Predictive Maintenance Dataset Data Set & 408.9239   & 409.5457  & 324.647   & 2.601  & 3.153  & 3.195   \\
Cloud                                             & 218.1206   & 217.8263  & 158.5564  & 2.942  & 3.473  & 6.804   \\ \hline
\end{tabular}}
\end{table}

According to Table \ref{tbl:rescomp6}, FFEAM and CE-AANN exhibit similar runtimes while AE demonstrates the fastest runtime. For instance, on the Baoan District traffic dataset, AE ran 25.7361 seconds faster than FFEAM. However, FFEAM's MAE is 193.075 lower than AE's, achieving a substantial improvement in filling accuracy. In general, the time difference between FFEAM and CE-AANN is insignificant and negligible. By contrast, compared to the conventional AE model, FFEAM comes with high model time consumption. However, its accuracy is substantially improved compared to the simple model. Therefore, the slightly longer time consumption of FFEAM is worthwhile since the model's increased accuracy justifies it.

\section{Conclusion}\label{s6.0}
A missing value filling model based on feature fusion enhanced autoencoder is proposed to address the key problems faced by the classical autoencoder model. The model constructed a new hidden layer of neural network by introducing two types of neurons, namely de-tracking neurons and radial basis function neurons, which combine the characteristics of the two types of neurons and exploit the interrelated features and common features of input data, so as to achieve multi-dimensional data feature fusion learning and improve the missing filling performance of the model. In addition, MVDC filling strategy was designed, and iterative training is optimized together with the model parameters to improve the filling perfromance. The experimental results on seven publicly available datasets and six artificial datasets show that FFEAM has better filling performance compared to the benchmark model. The current research has focused on methods to fill in missing numeric values. Non-numeric missing can generally be filled by plurality or by training a classifier to predict missing values of categorical variables, but how to effectively learn the implicit information within non-numeric and numeric attributes is a key issue to be addressed in future research.

\section*{Declaration of competing interest}
The authors declare that they have no known competing financial interests or personal relationships that could have appeared to influence the work reported in this paper.

\section*{Acknowledgements}
This work is supported by the National Key R\&D Program of China (No.2020AAA0105101), the National Natural Science Foundation of China (No.62276215, 62176221, 61976247)

\bibliography{sn-bibliography}


\begin{thebibliography}{30}
\ifx \bisbn   \undefined \def \bisbn  #1{ISBN #1}\fi
\ifx \binits  \undefined \def \binits#1{#1}\fi
\ifx \bauthor  \undefined \def \bauthor#1{#1}\fi
\ifx \batitle  \undefined \def \batitle#1{#1}\fi
\ifx \bjtitle  \undefined \def \bjtitle#1{#1}\fi
\ifx \bvolume  \undefined \def \bvolume#1{\textbf{#1}}\fi
\ifx \byear  \undefined \def \byear#1{#1}\fi
\ifx \bissue  \undefined \def \bissue#1{#1}\fi
\ifx \bfpage  \undefined \def \bfpage#1{#1}\fi
\ifx \blpage  \undefined \def \blpage #1{#1}\fi
\ifx \burl  \undefined \def \burl#1{\textsf{#1}}\fi
\ifx \doiurl  \undefined \def \doiurl#1{\url{https://doi.org/#1}}\fi
\ifx \betal  \undefined \def \betal{\textit{et al.}}\fi
\ifx \binstitute  \undefined \def \binstitute#1{#1}\fi
\ifx \binstitutionaled  \undefined \def \binstitutionaled#1{#1}\fi
\ifx \bctitle  \undefined \def \bctitle#1{#1}\fi
\ifx \beditor  \undefined \def \beditor#1{#1}\fi
\ifx \bpublisher  \undefined \def \bpublisher#1{#1}\fi
\ifx \bbtitle  \undefined \def \bbtitle#1{#1}\fi
\ifx \bedition  \undefined \def \bedition#1{#1}\fi
\ifx \bseriesno  \undefined \def \bseriesno#1{#1}\fi
\ifx \blocation  \undefined \def \blocation#1{#1}\fi
\ifx \bsertitle  \undefined \def \bsertitle#1{#1}\fi
\ifx \bsnm \undefined \def \bsnm#1{#1}\fi
\ifx \bsuffix \undefined \def \bsuffix#1{#1}\fi
\ifx \bparticle \undefined \def \bparticle#1{#1}\fi
\ifx \barticle \undefined \def \barticle#1{#1}\fi
\bibcommenthead
\ifx \bconfdate \undefined \def \bconfdate #1{#1}\fi
\ifx \botherref \undefined \def \botherref #1{#1}\fi
\ifx \url \undefined \def \url#1{\textsf{#1}}\fi
\ifx \bchapter \undefined \def \bchapter#1{#1}\fi
\ifx \bbook \undefined \def \bbook#1{#1}\fi
\ifx \bcomment \undefined \def \bcomment#1{#1}\fi
\ifx \oauthor \undefined \def \oauthor#1{#1}\fi
\ifx \citeauthoryear \undefined \def \citeauthoryear#1{#1}\fi
\ifx \endbibitem  \undefined \def \endbibitem {}\fi
\ifx \bconflocation  \undefined \def \bconflocation#1{#1}\fi
\ifx \arxivurl  \undefined \def \arxivurl#1{\textsf{#1}}\fi
\csname PreBibitemsHook\endcsname

\bibitem{bib1}
\begin{barticle}
\bauthor{\bsnm{Canbek}, \binits{G.}}:
\batitle{Gaining insights in datasets in the shade of “garbage in, garbage
  out” rationale: Feature space distribution fitting}.
\bjtitle{Wiley Interdisciplinary Reviews: Data Mining and Knowledge Discovery}
\bvolume{12}(\bissue{3}),
\bfpage{1456}
(\byear{2022})
\end{barticle}
\endbibitem

\bibitem{20}
\begin{barticle}
\bauthor{\bsnm{Xue}, \binits{Z.}},
\bauthor{\bsnm{Wang}, \binits{H.}}:
\batitle{Effective density-based clustering algorithms for incomplete data}.
\bjtitle{Big Data Mining and Analytics}
\bvolume{4}(\bissue{3}),
\bfpage{183}--\blpage{194}
(\byear{2021})
\end{barticle}
\endbibitem

\bibitem{bib3}
\begin{botherref}
\oauthor{\bsnm{Kabir}, \binits{S.}},
\oauthor{\bsnm{Farrokhvar}, \binits{L.}}:
Non-linear missing data imputation for healthcare data via index-aware
  autoencoders.
Health Care Management Science,
1--14
(2022)
\end{botherref}
\endbibitem

\bibitem{5}
\begin{barticle}
\bauthor{\bsnm{Lai}, \binits{X.}},
\bauthor{\bsnm{Wu}, \binits{X.}},
\bauthor{\bsnm{Zhang}, \binits{L.}},
\bauthor{\bsnm{Lu}, \binits{W.}},
\bauthor{\bsnm{Zhong}, \binits{C.}}:
\batitle{Imputations of missing values using a tracking-removed autoencoder
  trained with incomplete data}.
\bjtitle{Neurocomputing}
\bvolume{366},
\bfpage{54}--\blpage{65}
(\byear{2019})
\end{barticle}
\endbibitem

\bibitem{8}
\begin{bchapter}
\bauthor{\bsnm{Lai}, \binits{X.}},
\bauthor{\bsnm{Wu}, \binits{X.}},
\bauthor{\bsnm{Zhang}, \binits{L.}},
\bauthor{\bsnm{Zhang}, \binits{G.}}:
\bctitle{Imputation using a correlation-enhanced auto-associative neural
  network with dynamic processing of missing values}.
In: \bbtitle{International Symposium on Neural Networks},
pp. \bfpage{223}--\blpage{231}
(\byear{2019})
\end{bchapter}
\endbibitem

\bibitem{bib2}
\begin{botherref}
\oauthor{\bsnm{Liu}, \binits{K.}},
\oauthor{\bsnm{Lu}, \binits{N.}},
\oauthor{\bsnm{Wu}, \binits{F.}},
\oauthor{\bsnm{Zhang}, \binits{R.}},
\oauthor{\bsnm{Gao}, \binits{F.}}:
Model fusion and multiscale feature learning for fault diagnosis of industrial
  processes.
IEEE Transactions on Cybernetics
(2022)
\end{botherref}
\endbibitem

\bibitem{2}
\begin{barticle}
\bauthor{\bsnm{Vatanen}, \binits{T.}},
\bauthor{\bsnm{Osmala}, \binits{M.}},
\bauthor{\bsnm{Raiko}, \binits{T.}},
\bauthor{\bsnm{Lagus}, \binits{K.}},
\bauthor{\bsnm{Sysi-Aho}, \binits{M.}},
\bauthor{\bsnm{Ore{\v{s}}i{\v{c}}}, \binits{M.}},
\bauthor{\bsnm{Honkela}, \binits{T.}},
\bauthor{\bsnm{L{\"a}hdesm{\"a}ki}, \binits{H.}}:
\batitle{Self-organization and missing values in som and gtm}.
\bjtitle{Neurocomputing}
\bvolume{147},
\bfpage{60}--\blpage{70}
(\byear{2015})
\end{barticle}
\endbibitem

\bibitem{4}
\begin{bchapter}
\bauthor{\bsnm{Yousefi-Azar}, \binits{M.}},
\bauthor{\bsnm{Varadharajan}, \binits{V.}},
\bauthor{\bsnm{Hamey}, \binits{L.}},
\bauthor{\bsnm{Tupakula}, \binits{U.}}:
\bctitle{Autoencoder-based feature learning for cyber security applications}.
In: \bbtitle{2017 International Joint Conference on Neural Networks (IJCNN)},
pp. \bfpage{3854}--\blpage{3861}
(\byear{2017}).
\bcomment{IEEE}
\end{bchapter}
\endbibitem

\bibitem{6}
\begin{bchapter}
\bauthor{\bsnm{Daoud}, \binits{M.}},
\bauthor{\bsnm{Mayo}, \binits{M.}},
\bauthor{\bsnm{Cunningham}, \binits{S.J.}}:
\bctitle{Rbfa: radial basis function autoencoders}.
In: \bbtitle{2019 IEEE Congress on Evolutionary Computation (CEC)},
pp. \bfpage{2966}--\blpage{2973}
(\byear{2019}).
\bcomment{IEEE}
\end{bchapter}
\endbibitem

\bibitem{bib9}
\begin{barticle}
\bauthor{\bsnm{Ravi}, \binits{V.}},
\bauthor{\bsnm{Krishna}, \binits{M.}}:
\batitle{A new online data imputation method based on general regression auto
  associative neural network}.
\bjtitle{Neurocomputing}
\bvolume{138},
\bfpage{106}--\blpage{113}
(\byear{2014})
\end{barticle}
\endbibitem

\bibitem{19}
\begin{botherref}
\oauthor{\bsnm{LIU}, \binits{X.}},
\oauthor{\bsnm{DU}, \binits{S.}},
\oauthor{\bsnm{TENG}, \binits{F.}},
\oauthor{\bsnm{LI}, \binits{T.}}:
A missing value filling model based on feature fusion enhanced autoencoder.
in: 15th International FLINS Conferences on Machine learning, Multi agent and
  Cyber physical systems
(2022)
\end{botherref}
\endbibitem

\bibitem{bib4}
\begin{barticle}
\bauthor{\bsnm{Hamzah}, \binits{F.B.}},
\bauthor{\bsnm{Hamzah}, \binits{F.M.}},
\bauthor{\bsnm{Razali}, \binits{S.M.}},
\bauthor{\bsnm{Samad}, \binits{H.}}:
\batitle{A comparison of multiple imputation methods for recovering missing
  data in hydrological studies}.
\bjtitle{Civil Engineering Journal}
\bvolume{7}(\bissue{9}),
\bfpage{1608}--\blpage{1619}
(\byear{2021})
\end{barticle}
\endbibitem

\bibitem{9}
\begin{barticle}
\bauthor{\bsnm{Li}, \binits{D.}},
\bauthor{\bsnm{Zhang}, \binits{H.}},
\bauthor{\bsnm{Li}, \binits{T.}},
\bauthor{\bsnm{Bouras}, \binits{A.}},
\bauthor{\bsnm{Yu}, \binits{X.}},
\bauthor{\bsnm{Wang}, \binits{T.}}:
\batitle{Hybrid missing value imputation algorithms using fuzzy c-means and
  vaguely quantified rough set}.
\bjtitle{IEEE Transactions on Fuzzy Systems}
\bvolume{30}(\bissue{5}),
\bfpage{1396}--\blpage{1408}
(\byear{2021})
\end{barticle}
\endbibitem

\bibitem{10}
\begin{barticle}
\bauthor{\bsnm{Rumaling}, \binits{M.I.}},
\bauthor{\bsnm{Chee}, \binits{F.P.}},
\bauthor{\bsnm{Dayou}, \binits{J.}},
\bauthor{\bsnm{Chang}, \binits{J.}},
\bauthor{\bsnm{Sentian}, \binits{J.}}:
\batitle{Missing value imputation for pm10 concentration in sabah using nearest
  neighbour method (nnm) and expectation-maximization (em) algorithm}.
\bjtitle{Asian Journal of Atmospheric Environment}
\bvolume{14}(\bissue{1}),
\bfpage{62}--\blpage{72}
(\byear{2020})
\end{barticle}
\endbibitem

\bibitem{12}
\begin{botherref}
\oauthor{\bsnm{Ma}, \binits{B.}},
\oauthor{\bsnm{Li}, \binits{C.}},
\oauthor{\bsnm{Jiang}, \binits{L.}}:
A novel ground truth inference algorithm based on instance similarity for
  crowdsourcing learning.
Applied Intelligence,
1--13
(2022)
\end{botherref}
\endbibitem

\bibitem{13}
\begin{barticle}
\bauthor{\bsnm{Tutz}, \binits{G.}},
\bauthor{\bsnm{Ramzan}, \binits{S.}}:
\batitle{Improved methods for the imputation of missing data by nearest
  neighbor methods}.
\bjtitle{Computational Statistics \& Data Analysis}
\bvolume{90},
\bfpage{84}--\blpage{99}
(\byear{2015})
\end{barticle}
\endbibitem

\bibitem{14}
\begin{barticle}
\bauthor{\bsnm{Wang}, \binits{M.}},
\bauthor{\bsnm{Li}, \binits{D.}},
\bauthor{\bsnm{Qi}, \binits{K.}},
\bauthor{\bsnm{Xue}, \binits{C.}},
\bauthor{\bsnm{Yang}, \binits{E.}}:
\batitle{Sknn algorithm for filling missing oil data based on knn}.
\bjtitle{IOP Conference Series Materials Science and Engineering}
\bvolume{612},
\bfpage{032099}
(\byear{2019})
\end{barticle}
\endbibitem

\bibitem{15}
\begin{barticle}
\bauthor{\bsnm{Migdady}, \binits{H.}},
\bauthor{\bsnm{Al-Talib}, \binits{M.M.}}:
\batitle{An enhanced fuzzy k-means clustering with application to missing data
  imputation}.
\bjtitle{Electronic Journal of Applied Statistical Analysis}
\bvolume{11}(\bissue{2}),
\bfpage{674}--\blpage{686}
(\byear{2018})
\end{barticle}
\endbibitem

\bibitem{16}
\begin{barticle}
\bauthor{\bsnm{Li}, \binits{D.}},
\bauthor{\bsnm{Zhang}, \binits{H.}},
\bauthor{\bsnm{Li}, \binits{T.}},
\bauthor{\bsnm{Bouras}, \binits{A.}},
\bauthor{\bsnm{Yu}, \binits{X.}},
\bauthor{\bsnm{Wang}, \binits{T.}}:
\batitle{Hybrid missing value imputation algorithms using fuzzy c-means and
  vaguely quantified rough set}.
\bjtitle{IEEE Transactions on Fuzzy Systems}
\bvolume{PP},
\bfpage{1}--\blpage{1}
(\byear{2021})
\end{barticle}
\endbibitem

\bibitem{26}
\begin{barticle}
\bauthor{\bsnm{Deng}, \binits{W.}},
\bauthor{\bsnm{Guo}, \binits{Y.}},
\bauthor{\bsnm{Liu}, \binits{J.}},
\bauthor{\bsnm{Li}, \binits{Y.}},
\bauthor{\bsnm{Liu}, \binits{D.}},
\bauthor{\bsnm{Zhu}, \binits{L.}}:
\batitle{A missing power data filling method based on improved random forest
  algorithm}.
\bjtitle{Chinese Journal of Electrical Engineering}
\bvolume{5}(\bissue{4}),
\bfpage{33}--\blpage{39}
(\byear{2019})
\end{barticle}
\endbibitem

\bibitem{27}
\begin{bchapter}
\bauthor{\bsnm{Noei}, \binits{M.}},
\bauthor{\bsnm{Abadeh}, \binits{M.S.}}:
\bctitle{A genetic asexual reproduction optimization algorithm for imputing
  missing values}.
In: \bbtitle{2019 9th International Conference on Computer and Knowledge
  Engineering (ICCKE)},
pp. \bfpage{214}--\blpage{218}
(\byear{2019})
\end{bchapter}
\endbibitem

\bibitem{28}
\begin{barticle}
\bauthor{\bsnm{M.~Mostafa}, \binits{S.}},
\bauthor{\bsnm{S.~Eladimy}, \binits{A.}},
\bauthor{\bsnm{Hamad}, \binits{S.}},
\bauthor{\bsnm{Amano}, \binits{H.}}:
\batitle{Cbrl and cbrc: Novel algorithms for improving missing value imputation
  accuracy based on bayesian ridge regression}.
\bjtitle{Symmetry}
\bvolume{12}(\bissue{10}),
\bfpage{1594}
(\byear{2020})
\end{barticle}
\endbibitem

\bibitem{bib5}
\begin{barticle}
\bauthor{\bsnm{Tang}, \binits{S.}},
\bauthor{\bsnm{Yuan}, \binits{S.}},
\bauthor{\bsnm{Zhu}, \binits{Y.}}:
\batitle{Deep learning-based intelligent fault diagnosis methods toward
  rotating machinery}.
\bjtitle{Ieee Access}
\bvolume{8},
\bfpage{9335}--\blpage{9346}
(\byear{2019})
\end{barticle}
\endbibitem

\bibitem{21}
\begin{botherref}
\oauthor{\bsnm{Al-Kaabi}, \binits{K.}},
\oauthor{\bsnm{Monsefi}, \binits{R.}},
\oauthor{\bsnm{Zabihzadeh}, \binits{D.}}:
A framework to enhance generalization of deep metric learning methods using
  general discriminative feature learning and class adversarial neural
  networks.
Applied Intelligence,
1--19
(2022)
\end{botherref}
\endbibitem

\bibitem{bib6}
\begin{bchapter}
\bauthor{\bsnm{Saad}, \binits{M.}},
\bauthor{\bsnm{Chaudhary}, \binits{M.}},
\bauthor{\bsnm{Karray}, \binits{F.}},
\bauthor{\bsnm{Gaudet}, \binits{V.}}:
\bctitle{Machine learning based approaches for imputation in time series data
  and their impact on forecasting}.
In: \bbtitle{IEEE International Conference on Systems, Man, and Cybernetics
  (SMC)},
pp. \bfpage{2621}--\blpage{2627}
(\byear{2020})
\end{bchapter}
\endbibitem

\bibitem{25}
\begin{barticle}
\bauthor{\bsnm{Wang}, \binits{T.}},
\bauthor{\bsnm{Ke}, \binits{H.}},
\bauthor{\bsnm{Jolfaei}, \binits{A.}},
\bauthor{\bsnm{Wen}, \binits{S.}},
\bauthor{\bsnm{Haghighi}, \binits{M.S.}},
\bauthor{\bsnm{Huang}, \binits{S.}}:
\batitle{Missing value filling based on the collaboration of cloud and edge in
  artificial intelligence of things}.
\bjtitle{IEEE Transactions on Industrial Informatics}
\bvolume{18}(\bissue{8}),
\bfpage{5394}--\blpage{5402}
(\byear{2022})
\end{barticle}
\endbibitem

\bibitem{22}
\begin{barticle}
\bauthor{\bsnm{Sanjar}, \binits{K.}},
\bauthor{\bsnm{Bekhzod}, \binits{O.}},
\bauthor{\bsnm{Kim}, \binits{J.}},
\bauthor{\bsnm{Paul}, \binits{A.}},
\bauthor{\bsnm{Kim}, \binits{J.}}:
\batitle{Missing data imputation for geolocation-based price prediction using
  knn--mcf method}.
\bjtitle{ISPRS International Journal of Geo-Information}
\bvolume{9}(\bissue{4}),
\bfpage{227}
(\byear{2020})
\end{barticle}
\endbibitem

\bibitem{18}
\begin{barticle}
\bauthor{\bsnm{Zhao}, \binits{R.}},
\bauthor{\bsnm{Yan}, \binits{R.}},
\bauthor{\bsnm{Chen}, \binits{Z.}},
\bauthor{\bsnm{Mao}, \binits{K.}},
\bauthor{\bsnm{Wang}, \binits{P.}},
\bauthor{\bsnm{Gao}, \binits{R.X.}}:
\batitle{Deep learning and its applications to machine health monitoring}.
\bjtitle{Mechanical Systems and Signal Processing}
\bvolume{115},
\bfpage{213}--\blpage{237}
(\byear{2019})
\end{barticle}
\endbibitem

\bibitem{23}
\begin{barticle}
\bauthor{\bsnm{Lall}, \binits{R.}},
\bauthor{\bsnm{Robinson}, \binits{T.}}:
\batitle{The midas touch: Accurate and scalable missing-data imputation with
  deep learning}.
\bjtitle{Political Analysis}
\bvolume{30}(\bissue{2}),
\bfpage{179}--\blpage{196}
(\byear{2022})
\end{barticle}
\endbibitem

\bibitem{24}
\begin{bchapter}
\bauthor{\bsnm{Rodr\'{i}guez-Fdez}, \binits{I.}},
\bauthor{\bsnm{Canosa}, \binits{A.}},
\bauthor{\bsnm{Mucientes}, \binits{M.}},
\bauthor{\bsnm{Bugar\'{i}n}, \binits{A.}}:
\bctitle{{STAC}: a web platform for the comparison of algorithms using
  statistical tests}.
In: \bbtitle{Proceedings of the 2015 IEEE International Conference on Fuzzy
  Systems (FUZZ-IEEE)}
(\byear{2015})
\end{bchapter}
\endbibitem

\end{thebibliography}


\end{document}